\documentclass[]{fairmeta}

\usepackage{amsmath}
\usepackage{amssymb}
\usepackage{amsthm}

\usepackage{algorithm}
\usepackage{algorithmic}
\usepackage{sidecap}

\usepackage{wrapfig}

\theoremstyle{plain}
\newtheorem{theorem}{Theorem}[section]
\newtheorem{lemma}[theorem]{Lemma}

\theoremstyle{definition}
\newtheorem{definition}{Definition}[section]

\crefname{definition}{definition}{definitions}
\Crefname{definition}{Definition}{Definitions}

\crefname{property}{property}{properties}
\Crefname{property}{Property}{Properties}

\crefname{assumption}{assumption}{assumptions}
\Crefname{assumption}{Assumption}{Assumptions}
\Crefname{proposition}{Proposition}{Propositions}

\newtheorem{result}{Result}[section]
\Crefname{result}{Result}{Results}
\crefname{result}{Result}{Results}

\crefname{desideratum}{desideratum}{desiderata}
\Crefname{desideratum}{Desideratum}{Desiderata}

\title{A Scalable Measure of Loss Landscape Curvature for Analyzing the Training Dynamics of LLMs}

\author[1,2,*]{Dayal Singh Kalra}
\author[1]{Jean-Christophe Gagnon-Audet}
\author[1]{Andrey Gromov}
\author[1]{Ishita Mediratta}
\author[1]{Kelvin Niu}
\author[1]{Alexander H Miller}
\author[1]{Michael Shvartsman}

\affiliation[1]{Meta Superintelligence Labs}
\affiliation[2]{University of Maryland, College Park}

\contribution[*]{work done during an internship at Meta}

\abstract{
Understanding the curvature evolution of the loss landscape is fundamental to analyzing the training dynamics of neural networks. The most commonly studied measure, Hessian sharpness  ($\lambda_{\max}^H$) \textemdash the largest eigenvalue of the loss Hessian \textemdash determines local training stability and interacts with the learning rate throughout training. Despite its significance in analyzing training dynamics, direct measurement of Hessian sharpness remains prohibitive for Large Language Models (LLMs) due to high computational cost. We analyze \emph{critical sharpness} ($\lambda_c$), a computationally efficient measure requiring fewer than $10$ forward passes given the update direction $\Delta \bm{\theta}$. Critically, this measure captures well-documented Hessian sharpness phenomena, including progressive sharpening and Edge of Stability. Using this measure, we provide the first demonstration of these sharpness phenomena at scale, up to $7$B parameters, spanning both pre-training and mid-training of OLMo-2 models. We further introduce \emph{relative critical sharpness} ($\lambda_c^{1\to 2}$), which quantifies the curvature of one loss landscape while optimizing another, to analyze the transition from pre-training to fine-tuning and guide data mixing strategies. Critical sharpness provides practitioners with a practical tool for diagnosing curvature dynamics and informing data composition choices at scale. More broadly, our work shows that scalable curvature measures can provide actionable insights for large-scale training.
}

\date{\today}
\correspondence{\email{dayal@umd.edu}}

\metadata[Code]{\url{https://github.com/facebookresearch/scalable-curvature}}

\begin{document}

\maketitle

\section{Introduction}
\label{section:introduction}

{Understanding the evolution of the loss landscape $L(\bm{\theta})$ over the high-dimensional parameter space $\bm{\theta}$ is fundamental to analyzing the training dynamics of neural networks}. The loss landscape describes how the objective function changes with model parameters, and its geometry directly influences optimization, generalization, and stability throughout training~\citep{gilmer2022a,kalra2024why}. Intuitively, gradient-based optimization methods can swiftly navigate to the minima in smooth landscapes, while for rough landscapes, such methods can get trapped, hindering convergence or potentially converging to suboptimal solutions~\citep{losslandscapes}.

{The local geometry of the loss landscape is} commonly examined through the eigenvalues and eigenvectors of its Hessian matrix $H(\bm{\theta}):=\nabla_\theta^2 L(\bm{\theta})$~\citep{Wu2018_how_sgd,lewkowycz2020largelearningratephase,cohen2021gradient}. These eigenvalues provide insights into the local curvature of the loss landscape \textemdash large eigenvalues correspond to sharp, steep directions, while small eigenvalues indicate flat, smooth regions. 
The largest eigenvalue of the Hessian, which is often referred to as \emph{Hessian sharpness} $\lambda^H_{\max}$, quantifies the worst-case curvature of the landscape and is a fundamental metric in optimization. Its reciprocal, \emph{flatness} $1/\lambda_{\max}^H$, is a complementary measure used to describe the curvature.
In neural network optimization, Hessian sharpness relates to the local training stability~\citep{Wu2018_how_sgd,lewkowycz2020largelearningratephase}. For vanilla Gradient Descent (GD), if the learning rate exceeds the threshold $\sim 2/{\lambda_{\max}^H}$, loss increases as training `catapults' out of the local basin and eventually converges to a flatter region where the stability conditions are satisfied~\citep{lewkowycz2020largelearningratephase}. 
This observation generalizes to more complex optimization problems, including mini-batch settings and adaptive optimizers, albeit governed by different notions of sharpness~\citep{Wu2018_how_sgd,agarwala2025highdimensionalanalysisreveals,cohen2024adaptivegradientmethodsedge,kalra2024why}. For adaptive optimizers (e.g., Adam), the relevant measure is the pre-conditioned sharpness $\lambda^{PH}_{\max}$~\citep{cohen2024adaptivegradientmethodsedge}, which we formalize in \Cref{section:setup}.

\begin{wrapfigure}{r}{0.4\textwidth}
  \vspace{0pt}
  \centering
  \includegraphics[width=\linewidth]{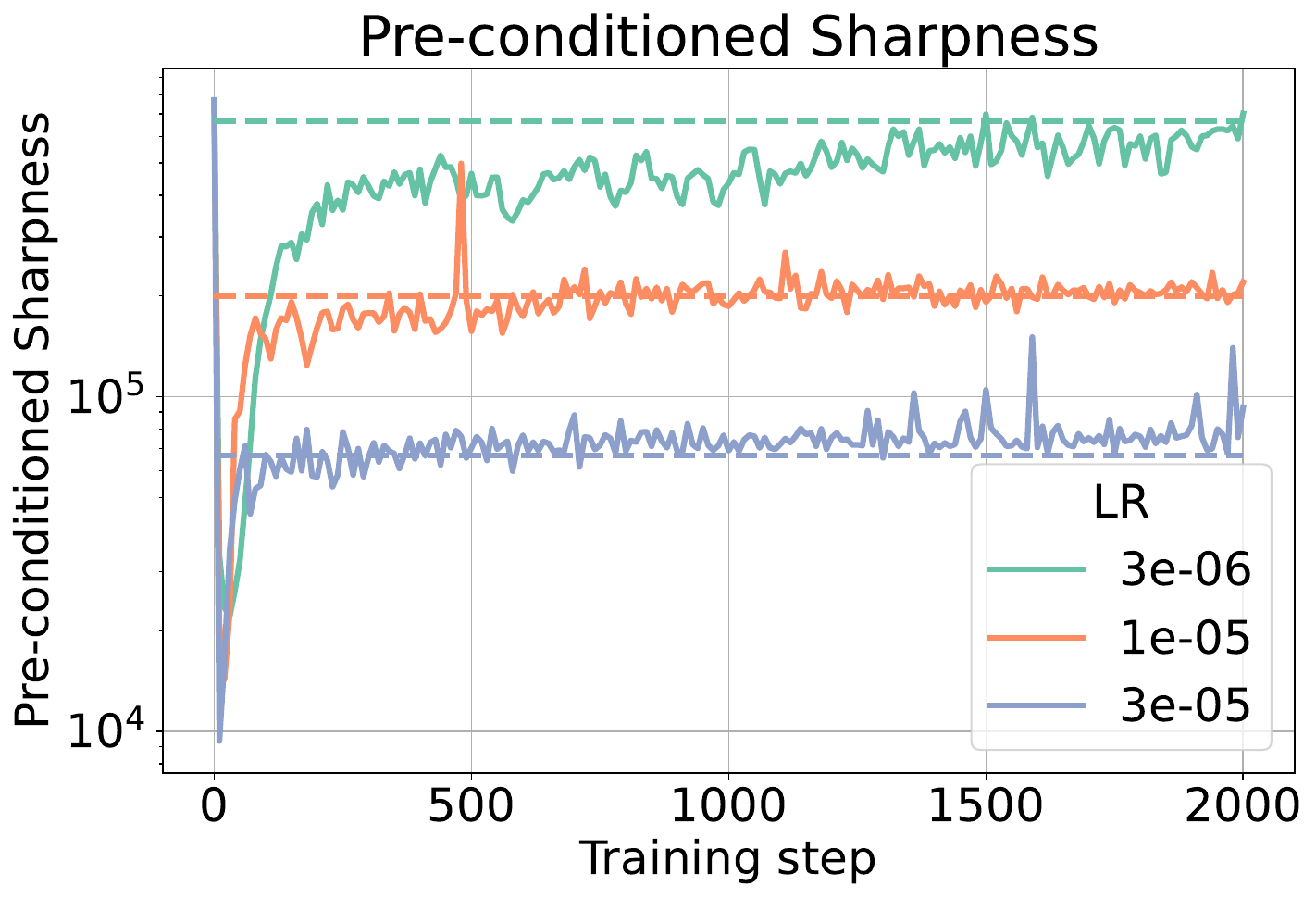}
  \caption{Pre-conditioned Hessian sharpness ($\lambda^{PH}_{\max}$) exhibits progressive sharpening and Edge of Stability (EoS) under constant learning rate. The dashed lines corresponding to the learning rate mark the EoS threshold.}
  \label{fig:intro_sharpness}
  \vspace{-10pt}
\end{wrapfigure}

As illustrated in \Cref{fig:intro_sharpness}, Hessian sharpness exhibits several robust trends throughout neural network training, particularly with constant learning rates. 
Training typically begins with an early reduction in Hessian sharpness~\citep{kalra2023phase,kalra2025universal} followed by a continuous increase until it reaches the stability threshold~\citep{Jastrzebski2020The,cohen2021gradient}. 
Once this threshold is reached, the training stabilizes through a self-stabilization mechanism~\citep{damian2023selfstabilization,cohen2025understanding} \textemdash instead of diverging as classical optimization would predict, Hessian sharpness begins to oscillate around this critical value. The continual increase in Hessian sharpness is referred to as \emph{progressive sharpening} while the subsequent oscillations around the critical threshold are termed the \emph{Edge of Stability} (EoS)~\citep{cohen2021gradient}.
For the more realistic setting of learning rate schedules involving warmup and decay, sharpness closely follows the learning rate schedule~\citep{gilmer2022a,kalra2024why,cohen2021gradient}. Due to its particularly close relationship with learning rate, sharpness can be used as a diagnostic tool for identifying training instabilities.

Beyond training dynamics, sharpness of the final solutions is linked to generalization properties, motivated by the intuition that flatter minima generalize better~\citep{Hochreiter1997FlatM}. However, this relationship has been called into question~\citep{pmlr-v187-kaur23a}, as empirical analyses show mixed results \textemdash
while flatter solutions generalize better in some settings, they can hurt performance in others. 
Sharpness also influences the early phase of training, with flatter initializations typically achieving better performance~\citep{NEURIPS2019_876e8108,kalra2024why}.

{Despite its value for examining model initialization, training dynamics, and generalization properties, analyzing sharpness at scale is challenging. 
Computing Hessian sharpness relies on iterative eigenvalue solvers (e.g., Power iteration, Lanczos), which require repeated Hessian-vector products (HVPs). However, HVP computation via automatic differentiation is often incompatible with modern training efficiency tools; kernels like Flash Attention \citep{dao2022flashattention} typically lack second-derivative implementations required for double backpropagation. Furthermore, iterative solvers can require hundreds of iterations with potential convergence failures.
As a result, most existing studies are restricted to small-scale experiments (typically $\sim 10$M parameters), leaving open questions about how sharpness evolves in Large Language Models (LLMs) at scale, and how it relates to optimization and downstream performance. In this work, we address this challenge by analyzing \emph{critical sharpness}, which leverages the relationship between curvature and training instability to serve as a computationally efficient proxy for the loss landscape curvature of LLMs.}

\subsection{Contributions}
Our contributions are as follows:
\begin{itemize}
    \item { We analyze critical sharpness, defined as $\lambda_c = 2 / \eta_c$, where $\eta_c$ is the \emph{critical learning rate}\textemdash the smallest learning rate that causes the training loss to increase in the next training step. To estimate the critical learning rate $\eta_c$, we perform an efficient line search along the update direction $\Delta \bm{\theta}$ from training. This procedure only requires forward passes, making it fully compatible with modern large-scale distributed training infrastructure, while avoiding the convergence issues of iterative eigenvalue solvers.
    In practice, we find that this procedure reliably estimates critical sharpness in only $5$-$6$ forward passes\footnote{except for the first iteration, which depends on the initial guess $\eta_0$ for line search.}}.
    \item \looseness -1
    We then examine the relationship between critical sharpness and Hessian sharpness. Under the quadratic loss approximation, we show that critical sharpness can be written as a weighted sum of Hessian eigenvalues and the two measures coincide when the gradient aligns with the largest eigenvector of the Hessian. We also generalize this result to adaptive optimizers.
    
    \item We demonstrate that critical sharpness reliably captures well-documented Hessian sharpness phenomena, such as \emph{progressive sharpening} and the \emph{Edge of Stability}~\citep{cohen2021gradient}. This makes critical sharpness a practical and computationally efficient proxy for sharpness. Leveraging OLMo-2 checkpoints~\citep{walsh2025} across pre-training and mid-training, we demonstrate progressive sharpening persists at realistic scales throughout training, including models with up to 7 billion parameters.

    \item We introduce a relative measure of critical sharpness $\lambda_c^{1 \to 2}$ to quantify the curvature of one loss relative to another, with application to analyzing the pre-training loss landscape during mid-training/fine-tuning. By varying the mix ratio of pre-training and fine-tuning (e.g. math), we show that increasing the proportion of pre-training data lowers relative critical sharpness, helping the model stay near the ``pre-trained basin'' during fine-tuning. We further demonstrate that downstream performance depends on basin retention: GSM8K~\citep{cobbe2021training} (math) benefits from leaving the pre-trained basin ($\eta > 2 / \lambda_c^{1 \to 2}$), while MMLU~\citep{hendryckstest2021} (generic reasoning) performs better when the basin is retained ($\eta < 2 / \lambda_c^{1 \to 2}$). This enables us to prescribe data mixing strategies to balance performance on math and generic reasoning tasks, depending on the desired outcome.

\end{itemize}

\section{Characterizing Critical Sharpness: Theory and Empirics}

\looseness -1
In this section, we characterize the dynamics of critical sharpness both theoretically and empirically. We first establish the theoretical relationship between critical sharpness and Hessian sharpness under a quadratic loss approximation, then validate these insights empirically on MLPs trained on CIFAR-10 using SGD across different batch sizes (\Cref{fig:sharpness_dynamics_cifar10_sgd}).

\subsection{Setup}
\label{section:setup}

Consider a model with trainable parameters $\bm{\theta} \in \mathbb{R}^n$, and let $L(\bm{\theta})$ denote the loss function with gradient $g(\bm{\theta}):=\nabla_\theta L(\bm{\theta})$ and Hessian $H(\bm{\theta}):=\nabla_\theta^2 L(\bm{\theta})$. Let $\{\lambda_{i}^H\}_{i = 1}^n$ and $\{{u}_{i}\}_{i = 1}^n$ denote the eigenvalues and eigenvectors of the Hessian, respectively. The largest eigenvalue $\lambda_{\max}^H = \max_i \lambda_{i}^H$, termed \emph{Hessian sharpness}, quantifies the worst-case local curvature. For adaptive optimizers with pre-conditioner $P(\bm{\theta})$ (e.g., Adam), the training dynamics is governed by the pre-conditioned Hessian $P(\bm{\theta})^{-1/2}H(\bm{\theta})P(\bm{\theta})^{-1/2}$, with eigenvalues $\{\lambda^{PH}_i \}_{i=1}^n$ and eigenvectors $\{{v}_i\}_{i=1}^n$.
We omit explicit $\bm{\theta}$ dependence when clear from context.

\subsection{Critical Sharpness: A Scalable Measure of Curvature}

\begin{figure*}[!t]
    \centering
    \includegraphics[width=0.35\linewidth]{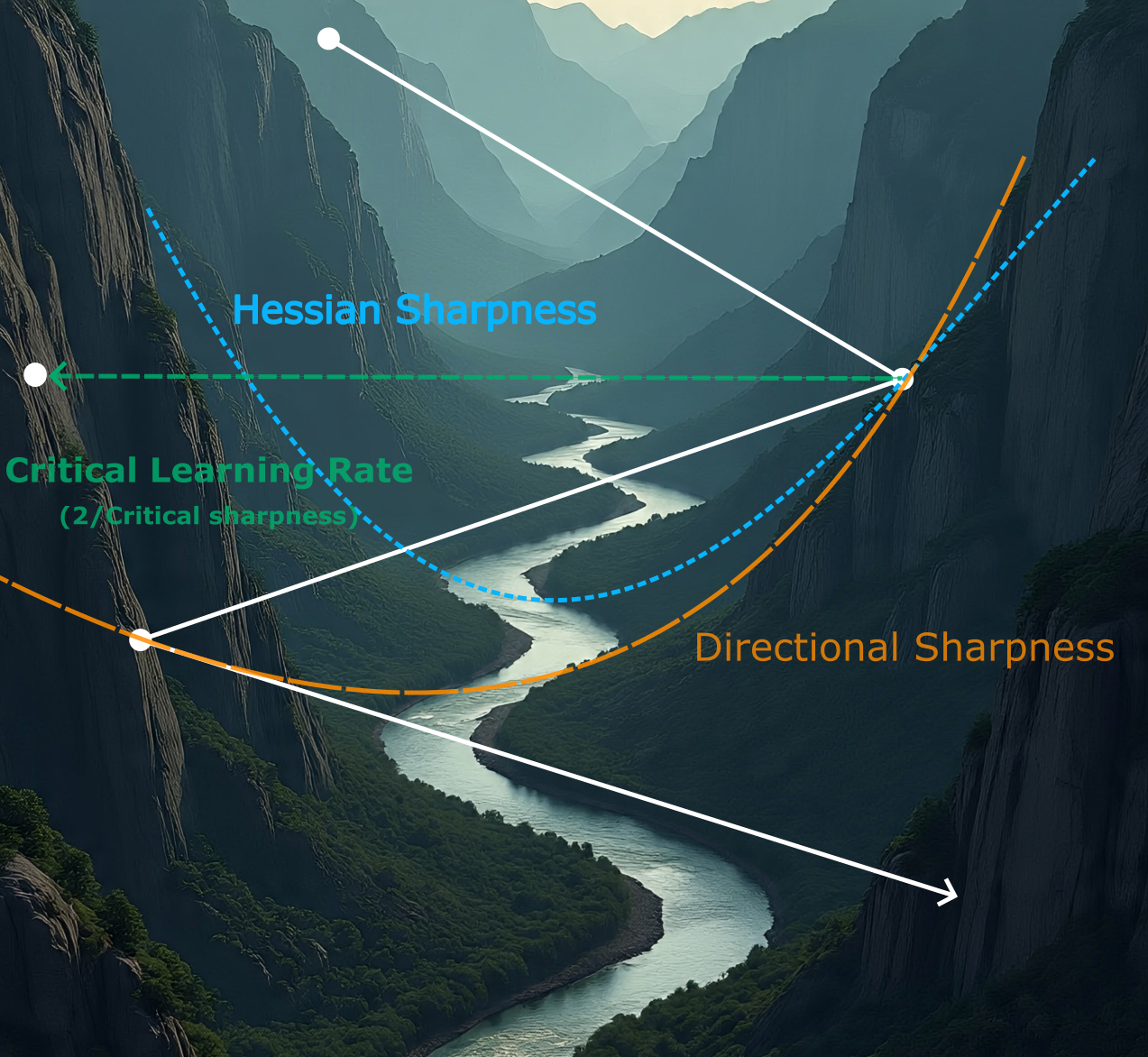}
    \caption{\emph{Comparison of different sharpness measures on an illustrative landscape featuring a sharp valley direction and a flat river direction} \protect\citep{wen2025understanding}. Hessian sharpness quantifies the curvature along the sharpest direction of the landscape (the valley). Directional sharpness measures the quadratic curvature along $\Delta \bm{\theta}$, while critical sharpness, its empirical counterpart, quantifies how far one can step along $\Delta \bm{\theta}$ before the loss increases.}
    \label{fig:landscape}
\end{figure*}

\begin{definition}[Critical learning rate $\eta_c$ and Sharpness $\lambda_c$]
\label{def:critical-lr-sharpness}   
Given an update direction $\Delta \bm{\theta}$, we define \emph{critical learning rate} as the smallest learning rate that causes the loss to increase in the next training step:
\begin{align}
    \eta_c = \inf \{ \eta > 0 \mid L(\bm{\theta} - \eta \Delta \bm{\theta} ) >  L(\bm{\theta})  \}.
\end{align}
Correspondingly, we define \emph{critical sharpness} as the scaled reciprocal of the critical learning rate $\lambda_c = 2 / \eta_c$.
\end{definition}
\paragraph{\textbf{Geometric Interpretation}:} 
Critical sharpness provides an intuitive, optimizer-aware measure of the local curvature of the loss landscape. Geometrically, it quantifies the ``natural length-scale'' of the landscape by estimating how far one can move in the current update direction without leaving the local region. While Hessian sharpness quantifies the curvature along the steepest direction of the landscape, critical sharpness measures the curvature along the update direction $\Delta \bm{\theta}$.
\Cref{fig:landscape} compares critical sharpness with Hessian sharpness on an illustrative landscape. 
\paragraph{\textbf{Efficient estimation of critical learning rate:}} To efficiently measure the critical learning rate, we build upon the line search method proposed by \citet{kalra2024why}. Given the update direction $\Delta \bm{\theta}$ from training, we compute the critical learning rate $\eta_c$ using a two-phase line search procedure:
\begin{enumerate}
    \item \emph{Exponential Search:} Starting from an initial guess $\eta_0$, we iteratively double (halve) $\eta$ until the loss $L(\bm{\theta} - \eta \Delta \bm{\theta})$ exceeds (or falls below) the current loss $L(\bm{\theta})$. This quickly identifies an interval $[\eta_{\text{lower}}, \eta_{\text{upper}}]$ containing $\eta_c$.
    \item \emph{Binary Search:} We then refine this interval using a binary search until the relative error $|1 - \eta_{\text{lower}}/ \eta_{\text{upper}}|$ falls below a threshold $\epsilon$. For $\epsilon = \frac{1}{2^k}$, the binary search converges in $k$ steps. 
\end{enumerate}
We approximate the critical learning rate using the mean of the final range $\eta_c \approx \frac{1}{2}(\eta_{\text{lower}} + \eta_{\text{upper}})$.
In practice, the exponential search typically requires only $1$–$2$ iterations (except for the first use, which depends on the initial guess $\eta_0$), and setting $\epsilon = 1/16$ (i.e., $k=4$) provides a reliable and efficient estimate of the critical learning rate in approximately $5$–$6$ forward passes, as shown in \Cref{fig:sharpness_dynamics_cifar10_sgd}.
Our approach is scalable, as it relies solely on forward passes to evaluate $L(\bm{\theta} - \eta \Delta \bm{\theta})$ and leverages the same computational primitives as standard first-order distributed training, avoiding the challenges of Hessian-based methods.
In \Cref{appendix:critical-lr-estimation}, we provide the detailed algorithm and further discuss the design choices.

\subsection{The Relationship between Critical Sharpness and Hessian Sharpness}

\looseness -1
Before investigating the empirical behavior of critical sharpness, we establish its theoretical relationship with Hessian sharpness. This connection provides a principled foundation for understanding what critical sharpness captures.
To this end, we consider the quadratic approximation of the loss function along $\Delta \bm{\theta}$:
\begin{align}
    L(\bm{\theta} - \eta \Delta \bm{\theta}) \approx L(\bm{\theta}) - \eta \Delta \bm{\theta}^T g(\bm{\theta}) + \frac{1}{2} \eta^2 \Delta \bm{\theta}^T H(\bm{\theta}) \Delta \bm{\theta}. \nonumber
\end{align}
Within this approximation, the loss will increase if the learning rate $\eta$ exceeds $2 / \lambda_{\text{dir}}$, where $\lambda_{\text{dir}}$ is the \emph{directional sharpness}~\citep{pan2022toward,roulet2024stepping}:
\begin{definition}[Directional Sharpness $\lambda_{\text{dir}}$]
The directional sharpness along the update direction $\Delta \bm{\theta}$ is:
\begin{align}
    \lambda_{\text{dir}} = \frac{\Delta \bm{\theta}^T H(\bm{\theta}) \Delta \bm{\theta} }{\Delta \bm{\theta}^T g(\bm{\theta})}.
\end{align}
\end{definition}

\begin{figure*}[t]
     \centering
     \begin{subfigure}[b]{0.32\textwidth}
         \centering
         \includegraphics[width = \linewidth]{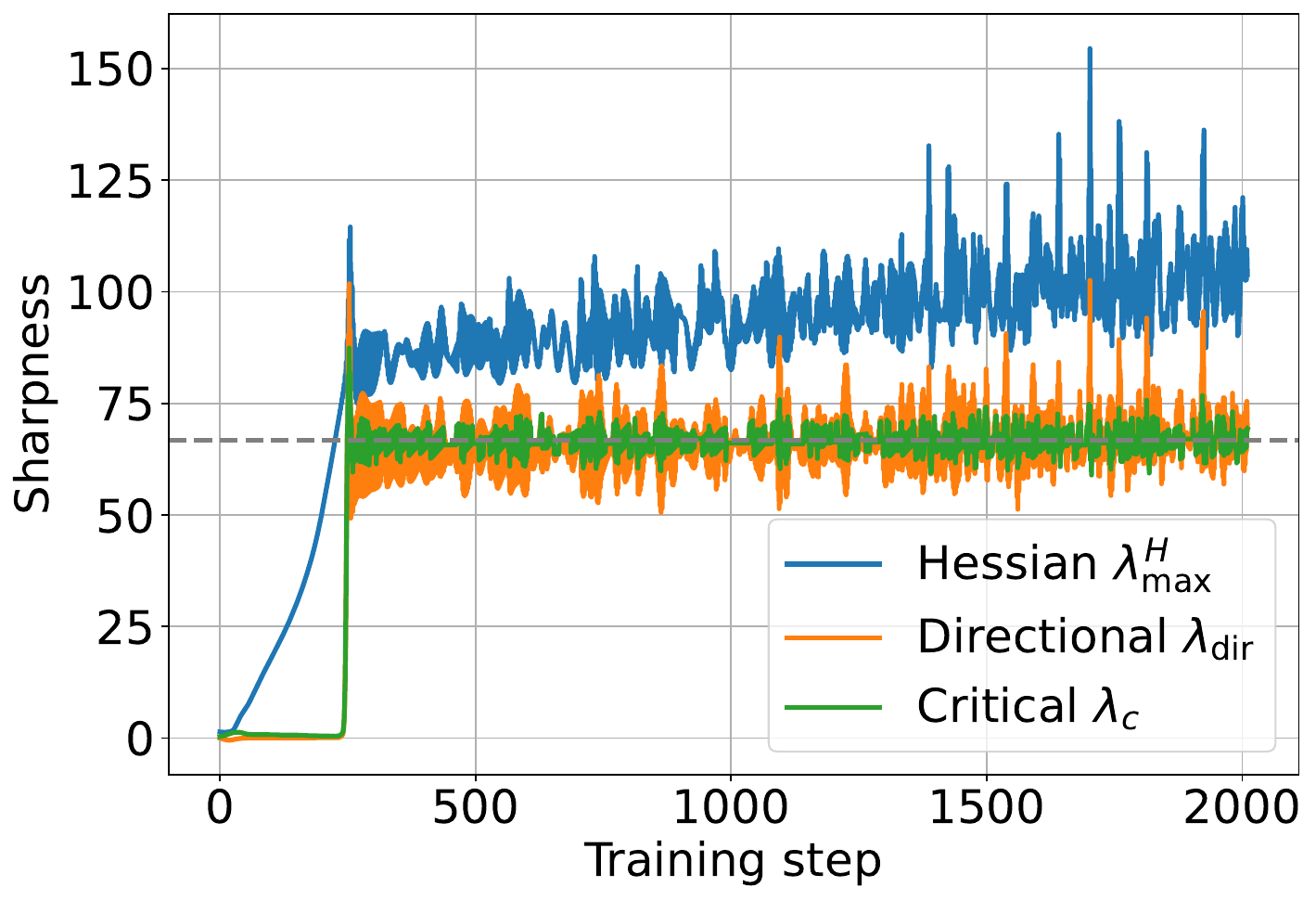}
         \caption{Batch size = 50,000 (GD)}
     \end{subfigure}
     \hfill
     \begin{subfigure}[b]{0.32\textwidth}
         \centering
         \includegraphics[width = \linewidth]{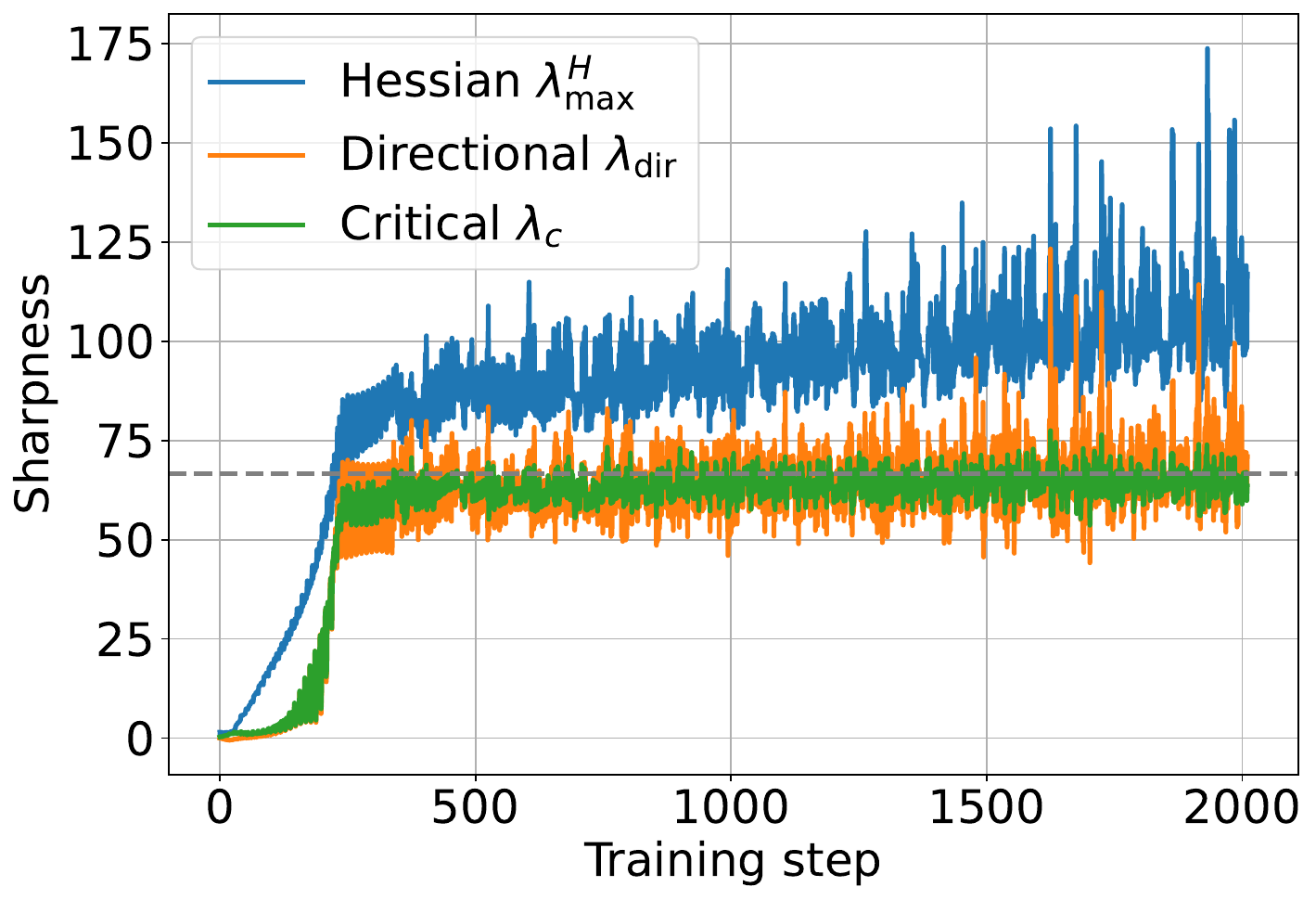}
         \caption{Batch size = 5000}
     \end{subfigure}
     \hfill
     \begin{subfigure}[b]{0.32\textwidth}
         \centering
         \includegraphics[width = \linewidth]{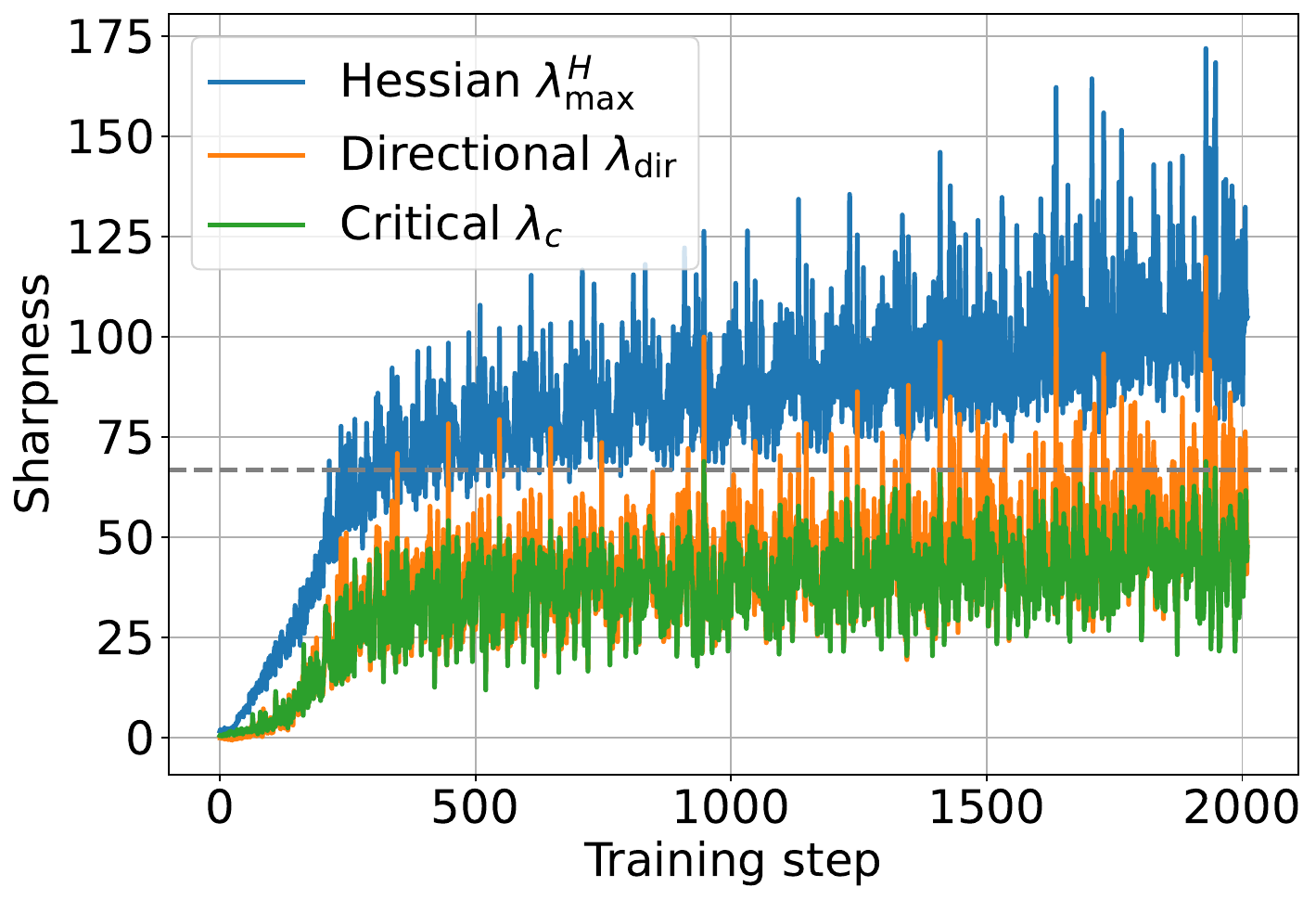}
        \caption{Batch size = 500}
     \end{subfigure}
     \caption{Comparison of different sharpness measures for MLPs trained on CIFAR-10 image classification task using SGD with learning rate $\eta$. 
     Both critical and directional sharpness exhibit progressive sharpening and Edge of Stability, albeit some deviations from Hessian sharpness.
     The dashed line denotes the Edge of Stability threshold, given by $2/\eta$.}
     \label{fig:sharpness_dynamics_cifar10_sgd}
\end{figure*}

Directional sharpness $\lambda_{\text{dir}}$ serves as an analytically tractable approximation to the empirically measured critical sharpness. \Cref{fig:sharpness_dynamics_cifar10_sgd} shows that directional sharpness closely tracks critical sharpness throughout training. 
This alignment suggests that the local quadratic approximation well captures the complex dynamics in this simplistic setting.
For Gradient Descent (GD), the directional sharpness can be expressed as a weighted sum of Hessian eigenvalues~\citep{roulet2024stepping}:
\begin{result}[Relationship between Directional and Hessian Sharpness for Gradient Descent]
\label{prop:dir_hessian_rel_gd}
\looseness -1
For Gradient Descent (GD), the directional sharpness $\lambda_{\text{dir}}$ can be expressed as a weighted sum of the Hessian eigenvalues $\{\lambda_i^H\}_{i=1}^n$, where the weights quantify the alignment of the gradient with Hessian eigendirections $\{u_i\}_{i=1}^n$:
\begin{align}
    \lambda_{\text{dir}} = \frac{\sum_{i = 1}^n c_i^2 \lambda_i^H}{\sum_{i = 1}^n c_i^2} \leq \lambda_{\max}^H,
\end{align}
where $c_i = \bm{g}^T \bm{u}_i$ is the projection of the gradient onto the $i^{th}$ eigenvector.
\end{result}

It follows that if the gradient $\bm{g}$ is perfectly aligned with the largest eigenvector, i.e., $ \bm{g} \propto \bm{u}_{\max}$, then directional sharpness coincides with Hessian sharpness. More generally, directional sharpness is always bounded above by Hessian sharpness. As a consequence, the gap between Hessian and directional sharpness quantifies the alignment of the gradient with the top eigendirection of the Hessian.
We now generalize this result to adaptive optimizers.

\begin{result}[Relationship between Directional and Hessian Sharpness for Adaptive optimizers]
\label{prop:dir_hessian_rel_adam}
For adaptive optimizers with pre-conditioner $P$ and update direction $P^{-1} \bm{g}$ (e.g., RMSProp), the directional sharpness $\lambda_{\text{dir}}$ can be expressed as a weighted sum of the pre-conditioned Hessian eigenvalues $\{\lambda^{PH}_i\}_{i=1}^n$, where the weights quantify the alignment of the pre-conditioned gradient $P^{-1/2} \bm{g}$ with pre-conditioned Hessian eigendirections $\{\bm{v}_i\}_{i=1}^n$:
\begin{align}
    \lambda_{\text{dir}} = \frac{\sum_{i = 1}^n c_i^2 \lambda_i^{PH}}{\sum_{i = 1}^n c_i^2} \leq \lambda_{\max}^{PH},
\end{align}
where $c_i = P^{-1/2}\bm{g}^T \bm{v}_i$ is the projection of the pre-conditioned gradient $P^{-1/2}g$ onto the $i^{th}$ eigenvector $\bm{v}_i$ of the pre-conditioned Hessian.
For momentum-based optimizers like Adam, this result serves as an approximation.
\end{result}
\looseness -1
Together, these results establish a connection between directional sharpness and Hessian sharpness, and indicate that the two measures diverge when the alignment between the gradient and the top eigendirection is small. In turn, Hessian and critical sharpness coincide when the loss surface is approximately locally quadratic, and the gradient is primarily along the top eigendirection. 
We detail the derivations of the above results in \Cref{appendix:directional_hessian_sharpness}.

We examine the differences between the three sharpness measures in \Cref{fig:sharpness_dynamics_cifar10_sgd}.
In the full-batch regime, Hessian sharpness exhibits progressive sharpening, eventually reaching the Edge of Stability (EoS) threshold. In contrast, both directional sharpness and critical sharpness remain nearly constant during the early stages of training, followed by an abrupt increase to the EoS threshold.
At smaller batch sizes, however, all three sharpness measures display a gradual increase from the onset of training. Notably, after crossing the EoS threshold, Hessian sharpness tends to oscillate above the threshold, whereas both critical sharpness and directional sharpness oscillate more closely around it.
Overall, critical sharpness exhibits progressive sharpening and Edge of Stability, albeit with some differences compared to Hessian sharpness discussed above. In the following section, we extend our analysis to more realistic, large-scale training settings.

\section{Critical Sharpness Dynamics at Scale}

\begin{figure*}[t]
     \centering
     \begin{subfigure}[b]{0.32\textwidth}
         \centering
         \includegraphics[width = \linewidth]{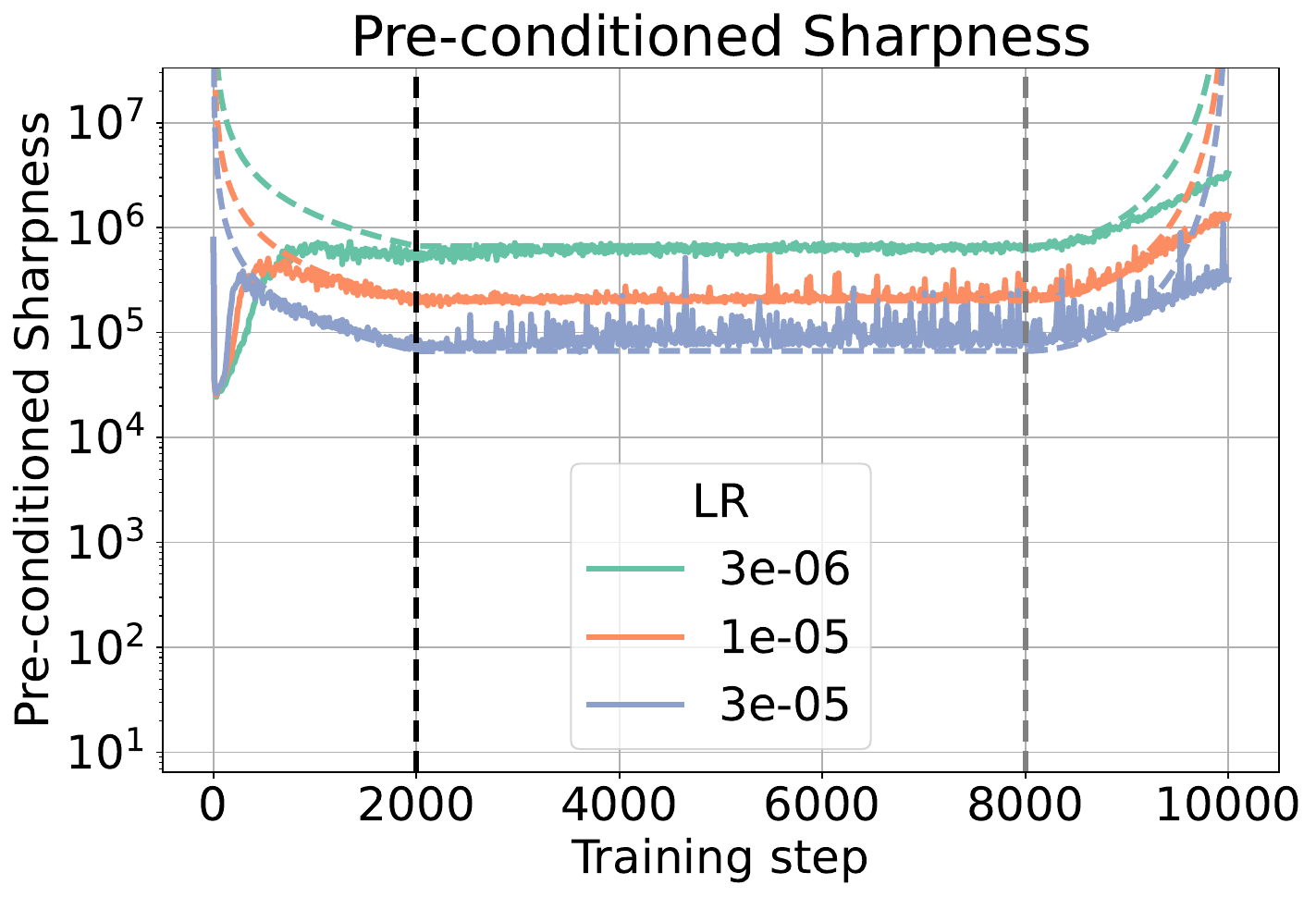}
         \caption{}
     \end{subfigure}
     \begin{subfigure}[b]{0.32\textwidth}
         \centering
         \includegraphics[width = \linewidth]{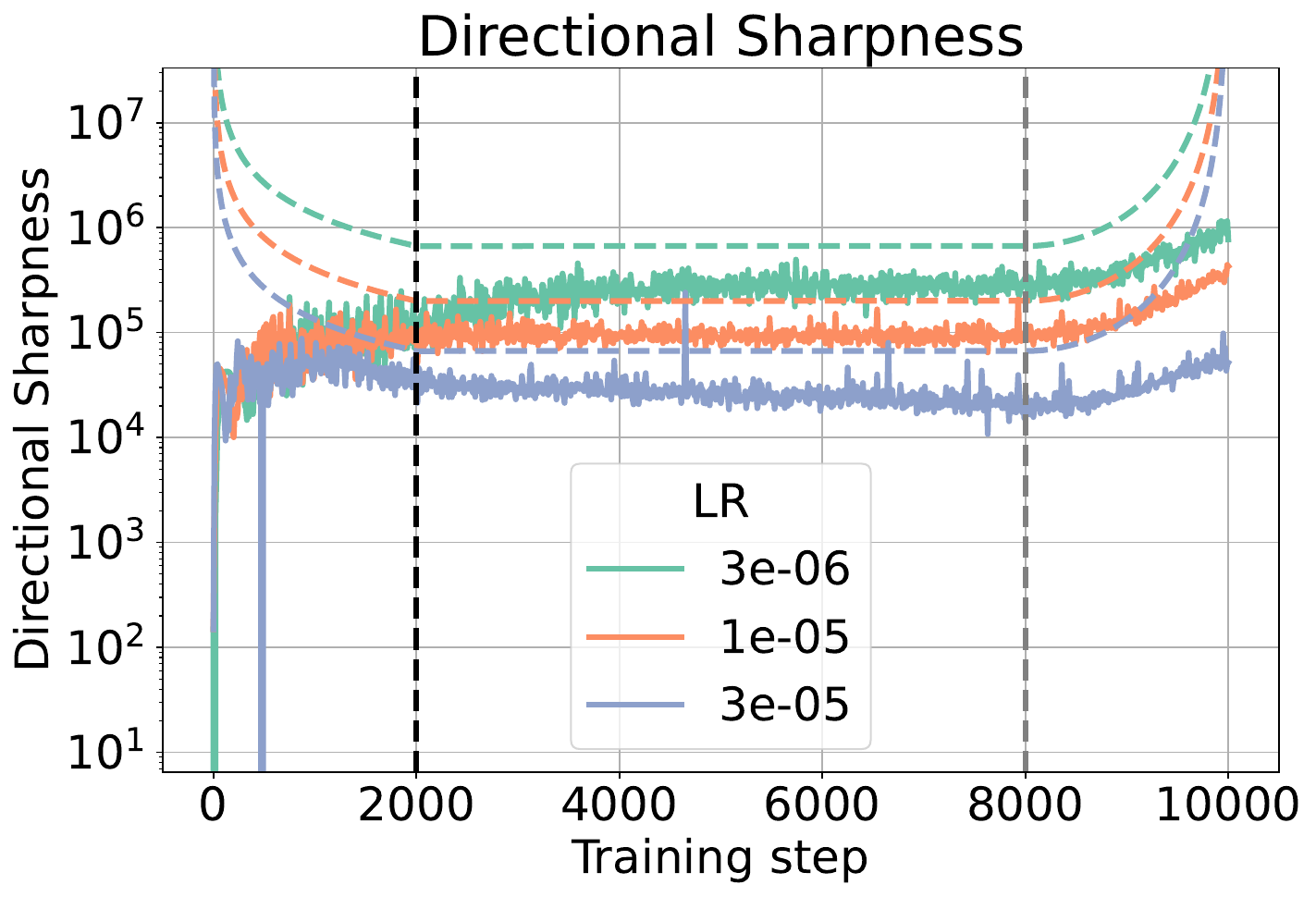}
         \caption{}
     \end{subfigure}
     \begin{subfigure}[b]{0.32\textwidth}
         \centering
         \includegraphics[width = \linewidth]{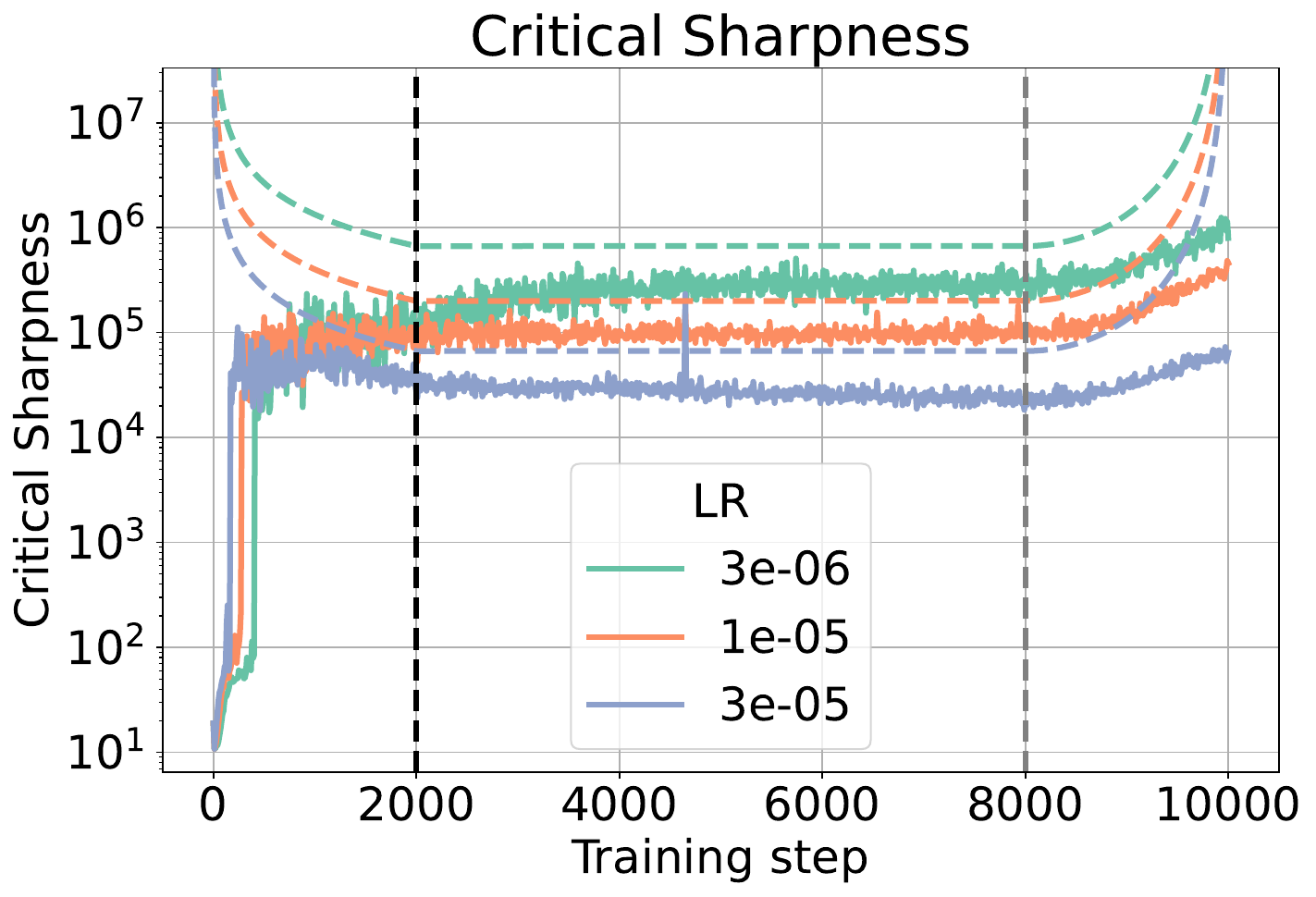}
         \caption{}
     \end{subfigure}
     \caption{Dynamics of pre-conditioned, directional, and critical sharpness during GPT-style Transformer training on FineWebEdu with AdamW using a Warmup-Stable-Decay (WSD) schedule. Critical sharpness tracks pre-conditioned sharpness throughout training, making it an effective proxy. The colored dashed lines denote the theoretical learning rate threshold $\left(\frac{2}{\eta} - \gamma \right)\left(\frac{1+\beta_1}{1-\beta
     _1} \right) $ and the two black vertical lines mark the end of warmup and stable phases.}
    \label{fig:sharpness_dynamics_nanogpt_pretraining}
\end{figure*}

Modern large-scale models are typically trained using Adam with weight decay~\citep{adamwloshchilov2018}, which helps mitigate training instabilities~\citep{dangelo2024why}. To analyze sharpness dynamics in this context, we first analyze the stability threshold for common optimizers with weight decay:
\begin{result}[Stability threshold for optimizers with weight decay]
For gradient descent and adaptive optimizers such as Adam, adding weight decay shifts the EoS threshold by a constant that depends on the decay strength $\gamma$:
\begin{align}
    \lambda^H_{\max} &= \frac{2}{\eta} - \gamma \quad & \text{(GD)} \nonumber \\
    \lambda^{PH}_{\max} &= \left(\frac{2}{\eta}-\gamma\right)\left( \frac{1+\beta_1}{1-\beta_1} \right) \quad & \text{(Adam)}
\end{align}
where $\eta$ is the learning rate, $\beta_1$ is Adam's momentum parameter, and $\gamma$ is the weight decay strength. 
We provide the detailed proofs in \Cref{appendix:wd_stability}.
\end{result}

\looseness -1
We now analyze GPT-style Transformers pre-trained for next-token prediction on the FineWebEdu dataset \citep{penedo2024the} using AdamW with Warmup-Stable-Decay (WSD) schedule~\citep{hu2024minicpm}.
\Cref{fig:sharpness_dynamics_nanogpt_pretraining} compares the dynamics of critical and pre-conditioned sharpness for three different learning rates. The pre-conditioned sharpness $\lambda^{PH}_{\max}$ exhibits progressive sharpening while following the learning rate schedule throughout training \textemdash it is pushed down during the warmup phase, stays constant at the stability threshold during the stable phase, and increases again when the learning rate is decayed. Critical sharpness and directional sharpness follow similar trends throughout training \textemdash they exhibit clear progressive sharpening and EoS behavior, while oscillating below the EoS threshold during the stable phase and capturing the increase in pre-conditioned sharpness during the decay phase.

\looseness -1
While progressive sharpening has been consistently documented at small scales, especially in full-batch settings, its manifestation and relevance at large scales, particularly in the context of online LLM training, remain largely unexplored. 
Having established that critical sharpness serves as an efficient and reliable proxy for progressive sharpening and Edge of Stability phenomena, we are now in the position to investigate whether progressive sharpening persists in large-scale LLM training.
To this end, we analyze the publicly available OLMo-2 $7$B checkpoints~\citep{walsh2025}. OLMo-2 provides checkpoints throughout both pre-training and mid-training stages, enabling us to study sharpness dynamics at scale without the computational cost of training such models from scratch.

\begin{figure*}[t]
     \centering
     \begin{subfigure}[b]{0.4\textwidth}
         \centering
         \includegraphics[width = \linewidth]{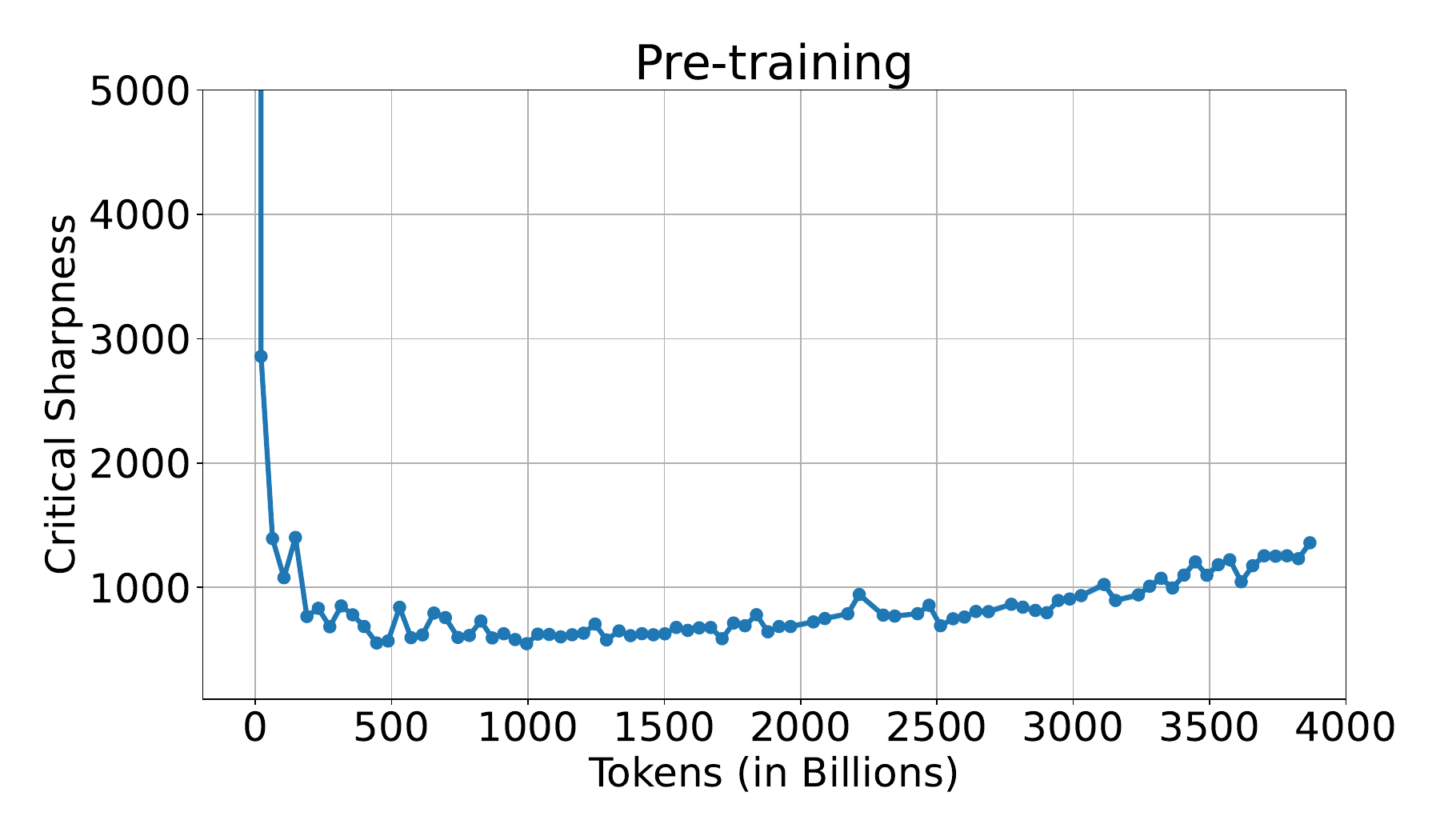}
         \caption{}
     \end{subfigure} 
     \begin{subfigure}[b]{0.4\textwidth}
         \centering
         \includegraphics[width = \linewidth]{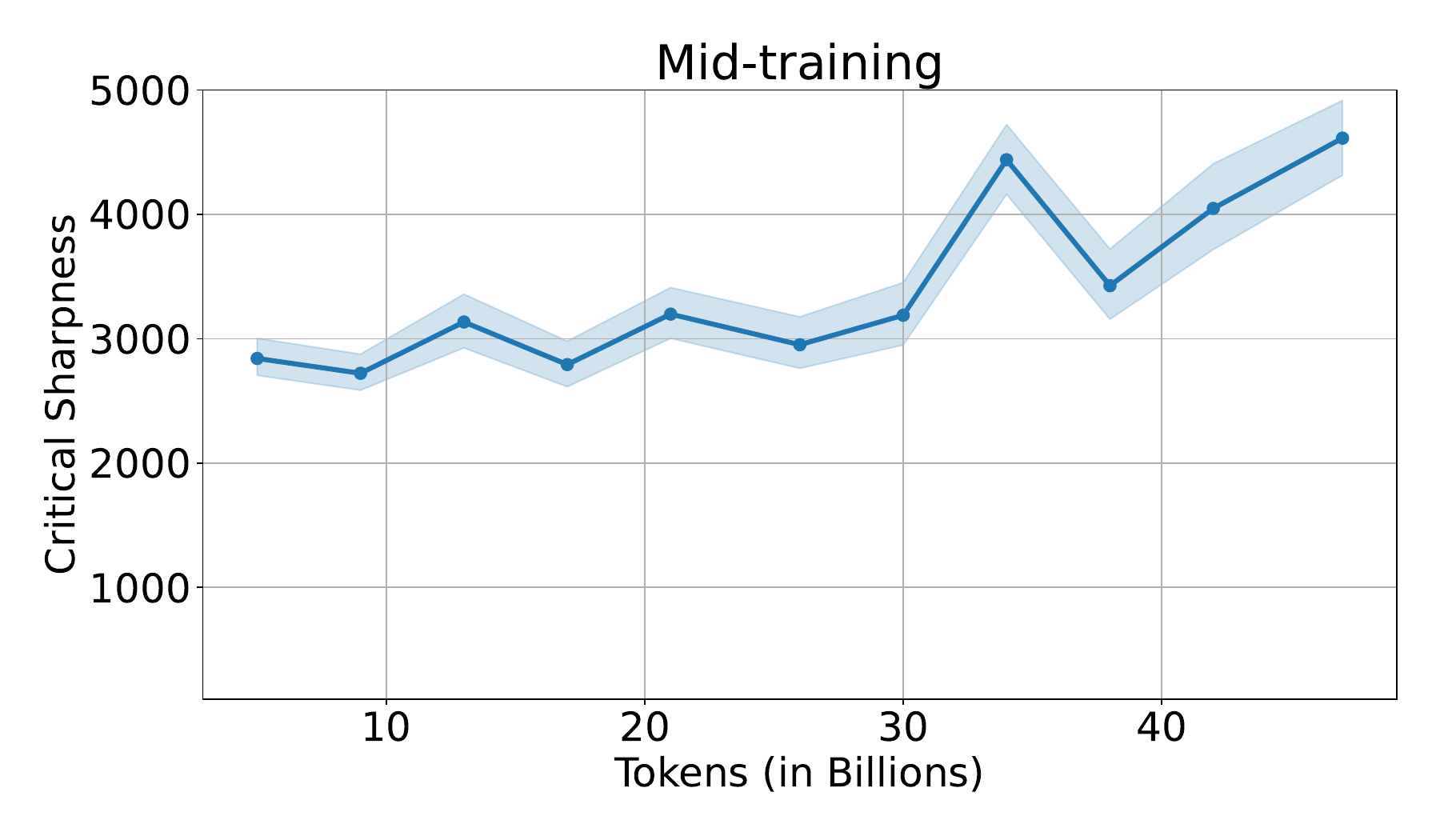}
         \caption{}
     \end{subfigure}
     \caption{Critical Sharpness of OLMo-2 $7$B exhibits progressive sharpening throughout pre-training and mid-training. In the mid-training figure, the band around the mean trend shows the deviation across the three runs. }
    \label{fig:crit_sharp_olmo_checkpoints}
\end{figure*}

\looseness -1
The OLMo-2 models are trained using a two-stage curriculum approach. During the pre-training stage ($>90\%$ of compute), the model is trained for $4$T tokens on the DCLM dataset~\citep{li2025datacomplmsearchgenerationtraining}, a diverse web-sourced corpus.
Following pre-training, the model undergoes a mid-training stage, where it is trained for $50$B tokens on a curated mix.
This mix includes academic papers, Wikipedia articles, instruction-following data, StackExchange documents, and approximately $50\%$ of the original pre-training dataset (DCLM).
During the pre-training stage, the model is trained using a learning rate schedule consisting of $2{,}000$ steps of warmup, followed by a cosine decay down to one-tenth of its peak value\footnote{For the OLMo-2 $7$B models, the cosine decay schedule is designed to reach one-tenth of the peak learning rate at $5$T tokens, but is truncated at $4$T tokens.}. By comparison, during the mid-training stage, the learning rate is linearly decayed to zero.
As the learning rate is continuously decreased throughout training after the warmup stage, the EoS threshold increases, and consequently we expect sharpness to progressively increase for the remainder of training.
To confirm this, we examine how the critical sharpness evolves across pre-training and mid-training using the publicly available checkpoints (see \Cref{appendix:experimental_details} for the details).
\Cref{fig:crit_sharp_olmo_checkpoints} shows that critical sharpness rapidly decreases during early training, but then continually increases (progressive sharpening) throughout both the pre-training and mid-training stages. This result provides the first empirical evidence of progressive sharpening at scale in practical LLM training settings.

\section{How much Pre-training data is needed to avoid Catastrophic Forgetting?}
\label{section:pre-training-mix-catastrophic-forgetting}

\looseness -1
To go beyond demonstrating pre-existing sharpness phenomena at scale, we next provide a new practical application of critical sharpness, focused on data mixing. We are motivated by the fact that during fine-tuning, neural networks are prone to \emph{catastrophic forgetting}, where the model performance degrades on the pretraining dataset and benchmarks as the model adapts to the new task~\citep{MCCLOSKEY1989109,lou2023_forgetting,mcleish2025teachingpretrainedlanguagemodels}. To mitigate this, several strategies have been proposed~\citep{Lange2022}, with mixing samples from the pre-training data being the most effective~\citep{ROBINS01061995}. This practice is reflected in large-scale training: for example, \citet{walsh2025} uses a mid-training mix consisting of approximately
$50\%$ pre-training data. Intuitively, we want to use as much new data as possible in fine-tuning, while using as little pre-training rehearsal data as possible to retain base capabilities. However, it remains unclear what fraction of pre-training data is sufficient to effectively prevent catastrophic forgetting.
To this end, we leverage critical sharpness to systematically examine the effect of adding pre-training data to the training mix during OLMo mid-training and provide actionable guidance for selecting the pre-training data fraction without exhaustive grid search.

The goal of mid-training or fine-tuning is to improve the performance on specialized domains or tasks while preserving generic capabilities acquired during pre-training. Intuitively, this requires the model to adapt to the new task, while staying within the ``pre-training basin'' \textemdash a region of the parameter space where the pre-training loss remains low.  Leaving this basin would be marked by an increase in the pre-training loss, and the critical learning rate quantifies exactly how far we can step before this occurs.
To formalize this intuition, we define relative critical learning rate and sharpness, as follows:
\begin{definition}[Relative Critical learning rate $\eta_c^{1 \to 2}$ and Sharpness $\lambda_c^{1 \to 2}$]
\label{def:relative-critical-lr-sharpness}   
Consider a model with parameters $\bm{\theta}$, two loss functions $L_1(\bm{\theta})$ and $L_2(\bm{\theta})$, and an update direction $\Delta \bm{\theta}$ derived from $L_2$. The \emph{relative critical learning rate} is the smallest learning rate for which taking a step in the direction $\Delta \bm{\theta}$ increases the loss $L_1(\bm{\theta})$:
\begin{align}
    \eta_c^{1\to2} = \inf \{ \eta > 0 \mid L_1(\bm{\theta} - \eta \Delta \bm{\theta} ) >  L_1(\bm{\theta})  \}.
\end{align}
The corresponding \emph{relative critical sharpness} is $\lambda_c^{1\to2} = 2 / \eta_c^{1\to2}$.

\end{definition}
This definition is general and applies whenever optimization on one objective may affect performance on another. The two losses can correspond to different tasks (e.g., pre-training vs. fine-tuning), different loss functions (e.g., next-token prediction vs. reinforcement learning objectives), or different data characteristics (e.g., short vs. long context). In this work, we focus on the fine-tuning setting, where $L_1({\bm{\theta}})$ is the pre-training loss on a general text corpus and $L_2({\bm{\theta}})$ is the fine-tuning loss, which often is a mixture of pre-training and specialized data (e.g., math).

To assess the impact of pre-training data fraction in the training mix during fine-tuning, we consider the OLMo-2 $7$B pre-trained checkpoint as our starting point. We define $L_2(\bm{\theta})$ as the loss on a mixture composed of DCLM (pre-training data) and the math subset of Dolmino mix ~\citep{walsh2025}, and compute the update direction $\Delta \bm{\theta}$ from this mixture. We then examine how the relative critical sharpness varies with the DCLM fraction in this mixture\footnote{When measuring the relative critical sharpness, we do not update the model parameters.}.

\begin{figure*}[t]
     \centering
     \begin{subfigure}[b]{0.31\textwidth}
         \centering
         \includegraphics[width = \linewidth]{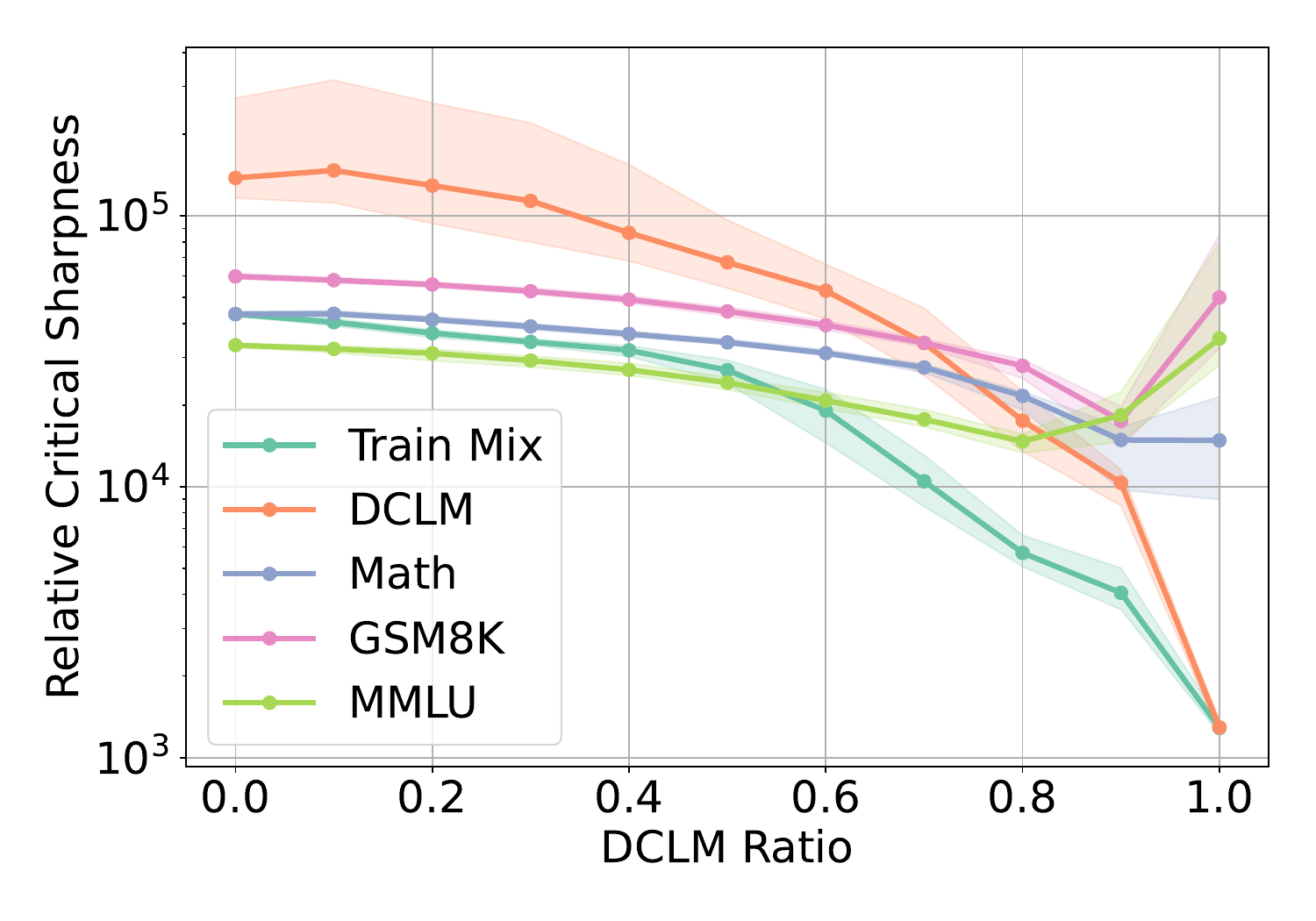}
         \caption{}
     \end{subfigure}
     \begin{subfigure}[b]{0.32\textwidth}
         \centering
         \includegraphics[width = \linewidth]{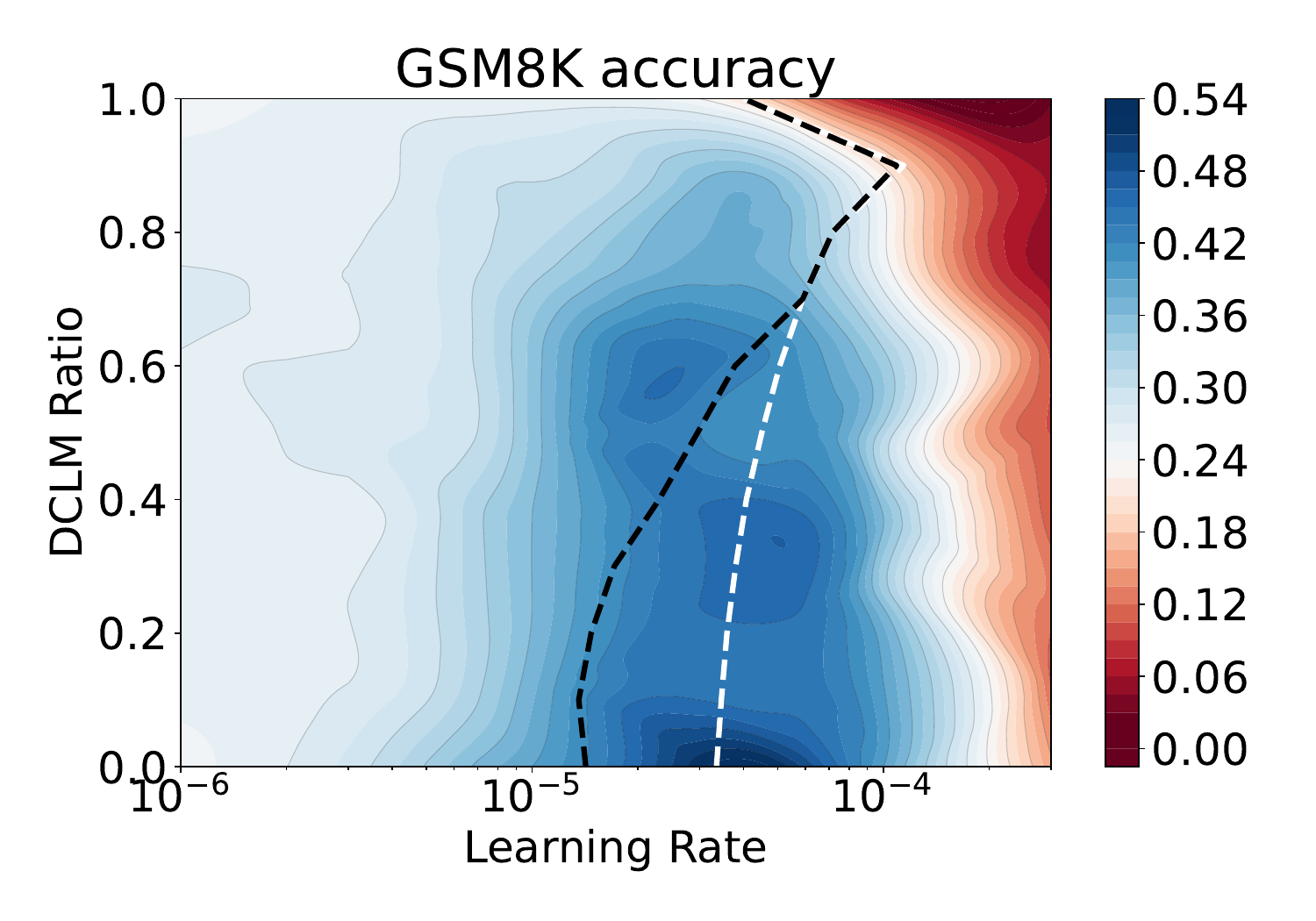}
         \caption{}
     \end{subfigure}
     \begin{subfigure}[b]{0.32\textwidth}
         \centering
         \includegraphics[width = \linewidth]{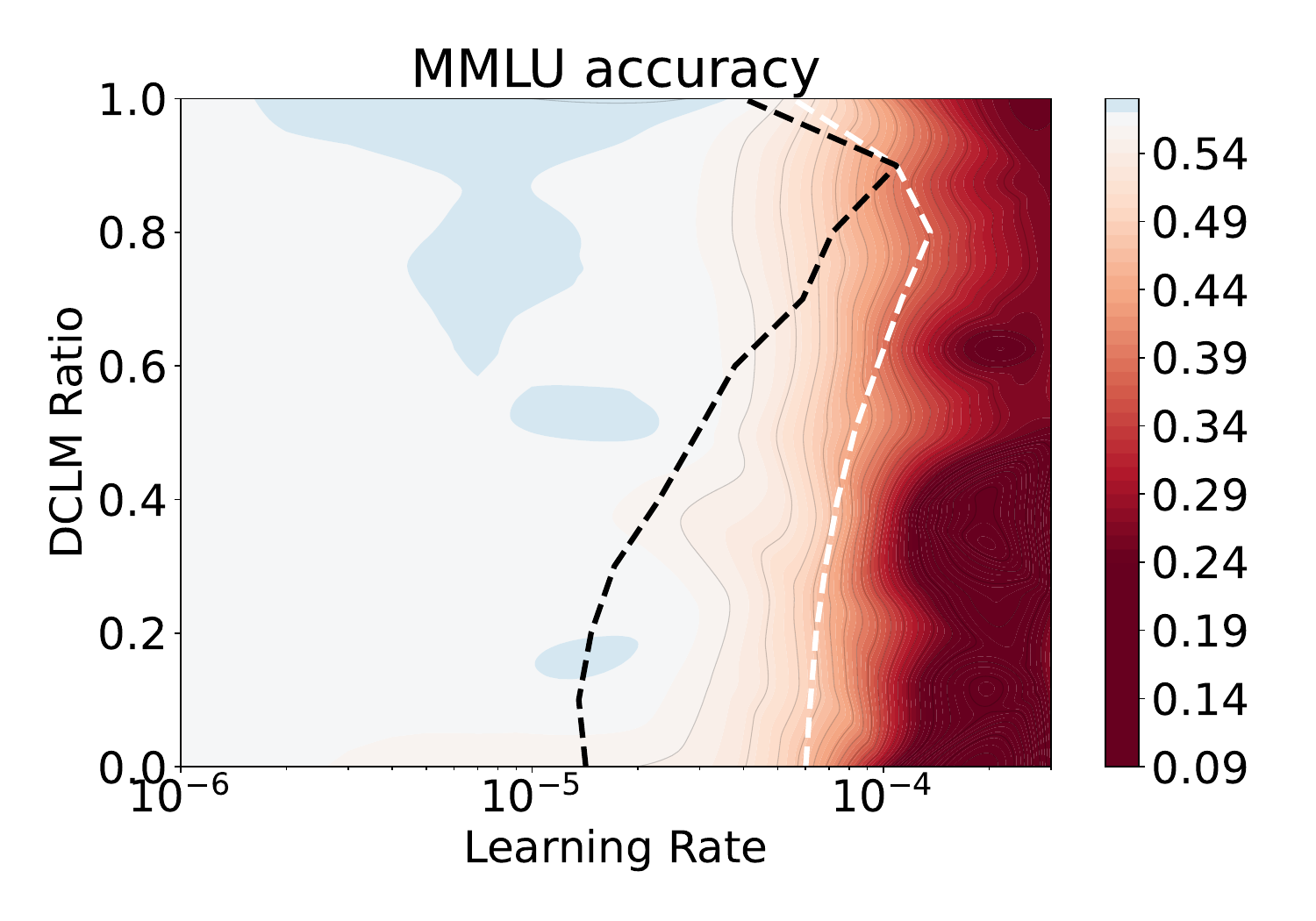}
         \caption{}
     \end{subfigure}
     \caption{(a) Relative critical sharpness (\Cref{def:relative-critical-lr-sharpness}) for various evaluation tasks ($L_1$). The shaded region around the mean trends denotes the variation across batches. (b, c) GSM8K and MMLU accuracy as a function of pre-training (DCLM) mix ratio and learning rate. Red indicates a decrease in performance relative to the checkpoint, white indicates no change, and blue indicates an improvement. The black dashed line denotes the smallest critical learning rate at the pre-trained checkpoint across tasks from (a), and the white dashed line denotes the critical learning rate for the corresponding task.}
    \label{fig:pretrain_mix}
\end{figure*}

\looseness -1
\Cref{fig:pretrain_mix}(a) shows how the relative critical sharpness varies with the DCLM ratio in the training mix for several evaluation tasks. When the mix contains mostly math data (low DCLM ratio), the sharpness for DCLM is an order of magnitude higher than for other tasks, indicating that the pre-training loss landscape is particularly sharp and sensitive to fine-tuning updates in this regime. As more pre-training data is added, the relative critical sharpness for most tasks decreases, suggesting that the landscapes align.
On the other hand, when the DCLM ratio approaches one, the sharpness for downstream tasks such as Math, GSM8K, and MMLU increases, meaning that these tasks become the limiting factor for the maximum stable learning rate. Notably, there is an intermediate DCLM ratio (around 0.7) where the sharpness curves for different tasks intersect, representing a sweet spot that allows for the largest possible learning rate without being constrained by any single task.

In \Cref{fig:pretrain_mix}(b, c), we evaluate the impact of the pre-training mix by training the pre-trained checkpoint for $1$B tokens and measuring downstream accuracy on GSM8K and MMLU benchmarks. In these plots, red indicates a decrease in performance relative to the pre-trained checkpoint, white indicates no change, and blue indicates an improvement. 
We observe a natural trade-off: GSM8K accuracy is typically maximized outside the pre-training basin (i.e., with low DCLM ratio and high learning rate), but this comes at the expense of MMLU performance, which is best preserved within the pre-training basin. Therefore, if the sole objective is to improve on the fine-tuning task (math), then training primarily on it with a large learning rate is effective, though it may lead to forgetting on other tasks (MMLU). In contrast, if the goal is to improve on the fine-tuning task (math) while maintaining pre-training performance (MMLU), it is essential to include pre-training data in the mix, with a sweet spot emerging around a DCLM ratio of $0.6$ and a learning rate of $3e$-$05$ in our experiment. Remarkably, this is close to the optimum of $ 0.7$ suggested by the critical sharpness analysis. This result highlights the importance of balancing task-specific and general data in the training mix. Once the pre-trained loss $L_1$ increases ($\eta > 2 / \lambda_c^{1 \to 2}$), there is no guarantee that further updates will improve $L_1$ again, unless the pre-training data is injected into the mix.

\looseness -1
In \Cref{appendix:relative_sharpness_olmo_midtraining}, we extend the relative critical sharpness analysis to the datamix used in OLMo-2 mid-training. The results are consistent: we observe a sweet spot at a DCLM ratio of
$0.6$, close to the $0.5$ ratio used in the original OLMo-2 training. We leave the validation of this prediction through downstream evaluation to future work.

\section{Related Works}

In neural network training, if the learning rate exceeds the threshold $\sim 2/\lambda^H_{\max}$, it causes the loss to increase in the next training step~\citep{Wu2018_how_sgd}. However, \citet{kalra2023phase} demonstrated that the empirical critical learning rate can be much higher than this theoretical threshold derived from convex analysis, with critical learning rate reaching up to $40/\lambda^H_{\max}$ when the sharpness naturally decreases during training.
Our work extends theirs by showing that a similarly large critical learning rate can also arise when sharpness increases during training, due to contributions from other eigendirections (\Cref{fig:sharpness_dynamics_cifar10_sgd}).
Beyond its role in characterizing training instabilities, Hessian sharpness also exhibits progressive sharpening during training. In particular, \citet{cohen2021gradient} showed that sharpness increases when the learning rate is decayed, which we corroborate with additional analysis at large scales (\Cref{fig:sharpness_dynamics_nanogpt_pretraining,fig:crit_sharp_olmo_checkpoints}).

The works most closely related to ours are those by \citet{kalra2024why}, \citet{roulet2024stepping} and \citet{bu2025gradient}. \citet{kalra2024why} used the critical learning rate to set the initial learning rate during warmup. By comparison, \citet{roulet2024stepping} proposed setting the learning rate at `edge' throughout training, i.e., $\eta_t = 2/\lambda_{\text{dir}}$, and show that it can outperform constant learning rate training, but not typical learning rate schedules, consisting of learning rate warmup followed by decay.~\citet{Vaswani2019Painless} also study setting the learning rate as $\eta = 1 / \lambda_{\text{dir}}$. 
\citet{bu2025gradient} proposed a generalized Newton method that uses forward passes to estimate local curvature for automatic learning rate selection.
In contrast, we use the critical learning rate to study the curvature dynamics of large-scale models and examine the effect of pre-training data on catastrophic forgetting.

\looseness -1
Catastrophic forgetting is a long-standing problem in neural networks, first identified by~\citet{MCCLOSKEY1989109}, where model performance degrades on previously learned tasks as it adapts to the new task. Approaches to mitigate catastrophic forgetting can be categorized into (i) regularization methods~\citep{Kirkpatrick2017,Ahn2019}, (ii) ensembling and parameter isolation~\citep{rusu2016progressive,Aljundi2017}, and  (iii) rehearsal~\citep{ROBINS01061995,Lopez-Paz2017}. Among these, rehearsal \textemdash mixing samples from the pre-training data \textemdash has become the most widely adopted strategy due to its simplicity and effectiveness. Recent works have demonstrated that catastrophic forgetting persists in LLM fine-tuning~
\citep{lou2023_forgetting,huang-etal-2024-mitigating,scialom-etal-2022-fine}. Particularly related to our work is that of~\citet{chen2025understandingpretrainingfinetuningloss}, who show that if fine-tuning retains previously learned capabilities, the model remains in ``most-case'' and ``worst-case'' basins. Theoretical works have further elucidated these dynamics; \citet{tahir2025features} argue that fine-tuning success relies on high feature overlap with the pre-trained model, while \citet{graldi2025the} show that staying in the ``lazy'' training regime can reduce forgetting. In the light of these works, relative critical sharpness can be used to quantify the feature overlap between the pre-trained model and the fine-tuning dataset and the critical learning rate required to stay in the lazy regime.

\section{Discussion and Conclusion}

In this work, we analyzed critical sharpness, a computationally efficient measure for studying the training dynamics of LLMs. Our results demonstrate that critical sharpness reliably captures key Hessian sharpness phenomena such as progressive sharpening and Edge of Stability, requiring fewer than $10$ forward passes, avoiding challenges associated with Hessian-based methods. Using this measure, we provided the first empirical evidence of progressive sharpening at the $7$B parameter scale.

We also introduced relative critical sharpness, which quantifies the curvature of one loss landscape along the update direction of another. Using this measure, we identified a sweet spot in the pre-training data fraction that balances specialization and retention during fine-tuning, enabling practitioners to evaluate data composition choices without extensive ablations. Beyond fine-tuning, relative critical sharpness provides a general framework for analyzing changes in the loss landscape due to distribution shifts, changes in loss functions, or modifications to the training data mixture.

We believe critical sharpness can extend beyond the settings studied here.
More broadly, our results demonstrate that scalable curvature measures can provide actionable insights for large-scale training, from understanding optimization dynamics to informing data composition decisions.

\section*{Acknowledgements}

We would like to thank Tianyu He and Darshil Doshi for helpful discussions and detailed comments on the manuscript, and Sean McLeish, Konstantin Mishchenko, Aaron Defazio, Maissam Barkeshli, Benjamin Therien, and Jesse Dodge for helpful discussions.
\bibliographystyle{assets/plainnat}
\bibliography{ref}

\clearpage
\newpage
\beginappendix

\section{Estimating Critical Learning Rate Using Forward Passes}
\label{appendix:critical-lr-estimation}

This section provides additional details on how to measure the critical learning rate using only forward passes. We generalize the line search method proposed by \citet{kalra2024why} to accommodate a generic initial guess $\eta_0$ and modify the exit condition for the binary search to get a better estimate of the critical sharpness.
Given the update direction $\Delta \bm{\theta}$ from training, we compute the critical learning rate $\eta_c$ using a two-phase line search procedure:

\paragraph{\textbf{Exponential Search:}} Starting from an initial guess $\eta_0$, we iteratively double (half) the learning rate $\eta$ until the loss $L(\bm{\theta} - \eta \Delta \bm{\theta})$ exceeds (or falls below) the current loss $L(\bm{\theta})$. This quickly identifies an interval $[\eta_{\text{lower}}, \eta_{\text{upper}}]$ containing $\eta_c$, with $\eta_{\text{upper}} = 2 \eta_{\text{lower}}$. Notably, each iteration requires only a single forward pass to evaluate the loss. The full algorithm is provided in \Cref{alg:exponential_search}.

\looseness -1
The number of exponential iterations depends on how close the initial guess $\eta_0$ is to the true critical learning rate $\eta_c$. After the first iteration, we update the initial guess to our current estimate of $\eta_c$, i.e., $\eta_0 = \eta_c$. In the subsequent steps, only $1-2$ exponential steps are needed, since $\eta_c$ tends to remain stable. However, if the training exhibits a large instability, causing the landscape to drastically change, more steps may be necessary.
To prevent indefinite iterations in such edge cases, we cap the exponential search to at most $40$ iterations, which corresponds to increasing or decreasing the learning rate by a factor of $10^{12}$ relative to the initial guess.

\begin{algorithm}[H]
\caption{Exponential Search}
\label{alg:exponential_search}
\begin{algorithmic}[1]
\STATE \textbf{Input:} Update direction $\Delta \bm{\theta}$, Initial loss $L(\bm{\theta})$, initial guess $\eta_0$, maximum iterations $N_{\max}$
\STATE \textbf{Output:} Interval $[\eta_{\text{lower}}, \eta_{\text{upper}}]$ containing the critical learning rate $\eta_c$
\STATE $\eta \gets \eta_0$
\STATE $i \gets 1 $ \hfill // Iteration
\STATE $L(\bm{\theta} - \eta \Delta \bm{\theta}) \gets \text{ComputeLoss}(\eta, \Delta \bm{\theta})$
\IF{$L(\bm{\theta} - \eta \Delta \bm{\theta}) < L(\bm{\theta})$}
    \STATE $\text{dir} \gets +1$ \hfill // Increase learning rate
\ELSE
    \STATE $\text{dir} \gets -1$ \hfill // Decrease learning rate
\ENDIF
\WHILE{$i < N_{\max}$}
    \STATE $\eta \gets \eta \times 2^{\text{dir}}$
    \STATE $L(\bm{\theta} - \eta \Delta \bm{\theta}) \gets \text{ComputeLoss}(\eta, \Delta \bm{\theta})$
    \STATE $i \gets i + 1$
    \IF{$\text{dir} = +1$ \AND $L(\bm{\theta} - \eta \Delta \bm{\theta}) > L(\bm{\theta})$}
        \STATE $\eta_{\text{lower}} \gets \eta / 2$
        \STATE $\eta_{\text{upper}} \gets \eta$
        \STATE \textbf{return} $[\eta_{\text{lower}}, \eta_{\text{upper}}]$
    \ENDIF
    \IF{$\text{dir} = -1$ \AND $L(\bm{\theta} - \eta \Delta \bm{\theta}) < L(\bm{\theta})$}
        \STATE $\eta_{\text{lower}} \gets \eta$
        \STATE $\eta_{\text{upper}} \gets 2\eta$
        \STATE \textbf{return} $[\eta_{\text{lower}}, \eta_{\text{upper}}]$
    \ENDIF
\ENDWHILE
\STATE \textbf{return} $[\eta, \eta]$
\end{algorithmic}
\end{algorithm}

\textbf{Binary Search:} We then refine this interval using a binary search. At each iterate, we evaluate the loss at the midpoint $\eta_{\text{mid}} = \frac{1}{2}(\eta_{\text{lower}} + \eta_{\text{upper}})$ and shorten the interval accordingly. This process is continued until the relative error $|1 - \eta_{\text{lower}}/ \eta_{\text{upper}}|$ falls below a specified threshold $\epsilon$. For $\epsilon = \frac{1}{2^k}$, the binary search converges in $k$ steps. In practice, we find that setting $\epsilon = 1/16$ (i.e., $k=4$) provides a reliable and efficient estimate of the critical learning rate.  The full algorithm is detailed in \Cref{alg:binary_search}.

\begin{algorithm}[H]
\caption{Binary Search}
\label{alg:binary_search}
\begin{algorithmic}[1]
\STATE \textbf{Input:} Update direction $\Delta \bm{\theta}$, Initial loss $L(\bm{\theta})$, initial interval $[\eta_{\text{lower}}, \eta_{\text{upper}}]$, tolerance $\epsilon$
\STATE \textbf{Output:} Interval $[\eta_{\text{lower}}, \eta_{\text{upper}}]$ containing the critical learning rate $\eta_c$ s.t. $\left| 1 - \frac{\eta_{\text{lower}}}{\eta_{\text{upper}}} \right| < \epsilon$
\STATE $\text{i} \gets 0$ \hfill // iteration
\WHILE{$\left | 1 - \eta_{\text{lower}} / \eta_{\text{upper}} \right| > \epsilon$}
    \STATE $\eta_{\text{mid}} \gets \frac{1}{2} (\eta_{\text{lower}} + \eta_{\text{upper}})$
    \STATE $L(\bm{\theta} - \eta_{\text{mid}} \Delta \bm{\theta}) \gets \text{ComputeLoss}(\eta_{\text{mid}}, \Delta \bm{\theta})$
    \STATE $i \gets i + 1$
    \IF{$L(\bm{\theta} - \eta_{\text{mid}} \Delta \bm{\theta}) > L(\bm{\theta})$}
        \STATE $\eta_{\text{upper}} \gets \eta_{\text{mid}}$
    \ELSE
        \STATE $\eta_{\text{lower}} \gets \eta_{\text{mid}}$
    \ENDIF
\ENDWHILE
\STATE \textbf{return} $[\eta_{\text{lower}}, \eta_{\text{upper}}]$
\end{algorithmic}
\end{algorithm}

Finally, we approximate the critical learning rate using the mean of the final range $\eta_c \approx \frac{1}{2}(\eta_{\text{lower}} + \eta_{\text{upper}})$. Overall, this procedure reliably computes the critical learning rate in $5-6$ forward passes.

\section{Experimental Details}
\label{appendix:experimental_details}

\textbf{\Cref{fig:intro_sharpness}:} We considered GPT-style Pre-LN Transformers consisting of $12$ layers, with an embedding dimension of $n_{\text{embd}} = 768$. The model in total has $\sim 100$M parameters. We train the model on the $10$B token subset of the FineWebEdu dataset with $\sim 1$M tokens per step using AdamW with constant learning rate and hyperparameters $\beta_1 = 0.9$ and $\beta_2 = 0.95$. 

\textbf{\Cref{fig:sharpness_dynamics_cifar10_sgd}}: We consider Multi-Layer Perceptrons (MLPs) consisting of four layers, with width $512$ and GeLU activation function. The models were trained on the CIFAR-10 image classification task using SGD with a constant learning rate of $\eta = $3e-02. We experimented with three different batch sizes: $B \in [500, 5000, 50000]$, with a batch size of $50,000$ corresponds to full-batch gradient descent.

\textbf{\Cref{fig:sharpness_dynamics_nanogpt_pretraining}}: We considered GPT-style Pre-LN Transformers consisting of $12$ layers, with an embedding dimension of $n_{\text{embd}} = 768$. The model in total has $\sim 100$M parameters. We train the model on the $10$B token subset of FineWebEdu dataset with $\sim 1$M tokens per step using AdamW and hyperparameters $\beta_1 = 0.9$ and $\beta_2 = 0.95$.  

\textbf{\Cref{fig:crit_sharp_olmo_checkpoints}:} We evaluate the publicly available OLMo-2 7B checkpoint at both the pre-training and mid-training stages. For each checkpoint, we estimate the critical learning rate using the AdamW optimizer with a batch size of $16$ and hyperparameters $\beta_1 = 0.9$ and $\beta_2 = 0.99$. 

 \paragraph{Methodological Note:} We acknowledge two necessary approximations. First, the original training utilizes a large batch size of approximately $4M$ tokens, whereas our analysis uses a relatively small ($60\times$ smaller) batch size of $0.065$M tokens for computational tractability. While this alters the EoS threshold, this does not hinder us from observing progressive sharpening.
Second, since we do not have access to OLMo’s original optimizer states, we first accumulate Adam’s first and second moments over a warmup period of $100$ steps, which we find to be sufficient for the second moment to stabilize. After this warm-up period, we measure the critical learning rate for the next $100$ steps and report the average value across these steps. For the pre-training checkpoints, we evaluate the model using the DCLM dataset, whereas the mid-training checkpoints are evaluated using the Dolmino mix dataset~\citep{walsh2025}.
Importantly, model parameters are not updated during this experiment.

\textbf{\Cref{fig:pretrain_mix}:} We consider the last OLMo-2 $7$B as our starting point and further train on a mixture composed of DCLM and the math subset of Dolmino mix~\citep{walsh2025} with a context length of $4096$. In \Cref{fig:pretrain_mix}(a), we evaluate the relative critical sharpness for several evaluation tasks across DCLM ratios. (b, c) We further train the pre-trained checkpoint on the train mix for $16,000$ steps ($1$B tokens) using AdamW with a constant learning rate \footnote{We consider a constant learning rate because we are only training the model for $1$B tokens, as compared to the OLMo mid-training for $50$B tokens.}, batch size of $16$ and hyperparameters $\beta_1 = 0.9$ and $\beta_2 = 0.99$

\begin{figure*}[!h]
     \centering
     \begin{subfigure}[b]{0.32\textwidth}
         \centering
         \includegraphics[width = \linewidth]{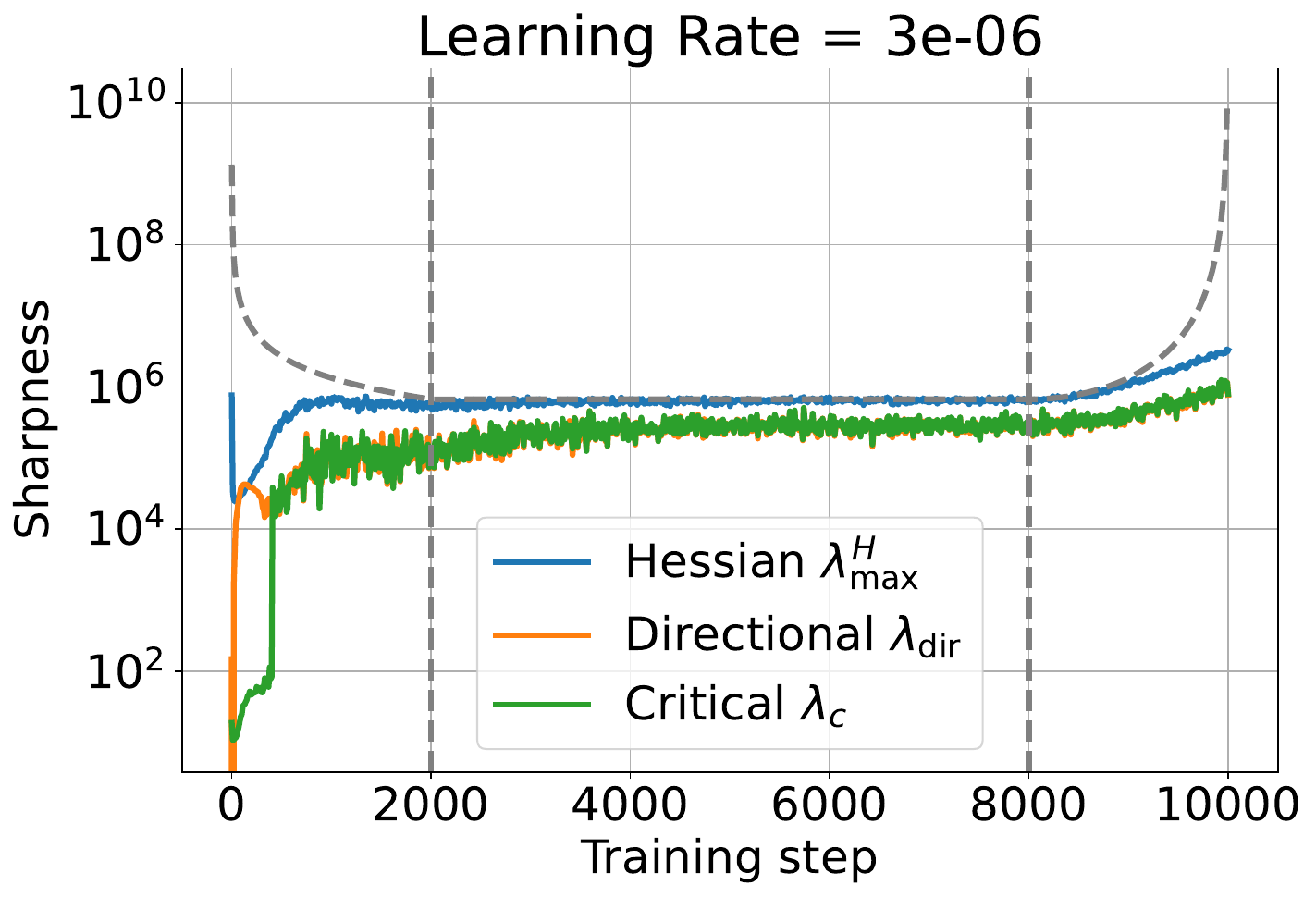}
         \caption{}
     \end{subfigure}
     \begin{subfigure}[b]{0.32\textwidth}
         \centering
         \includegraphics[width = \linewidth]{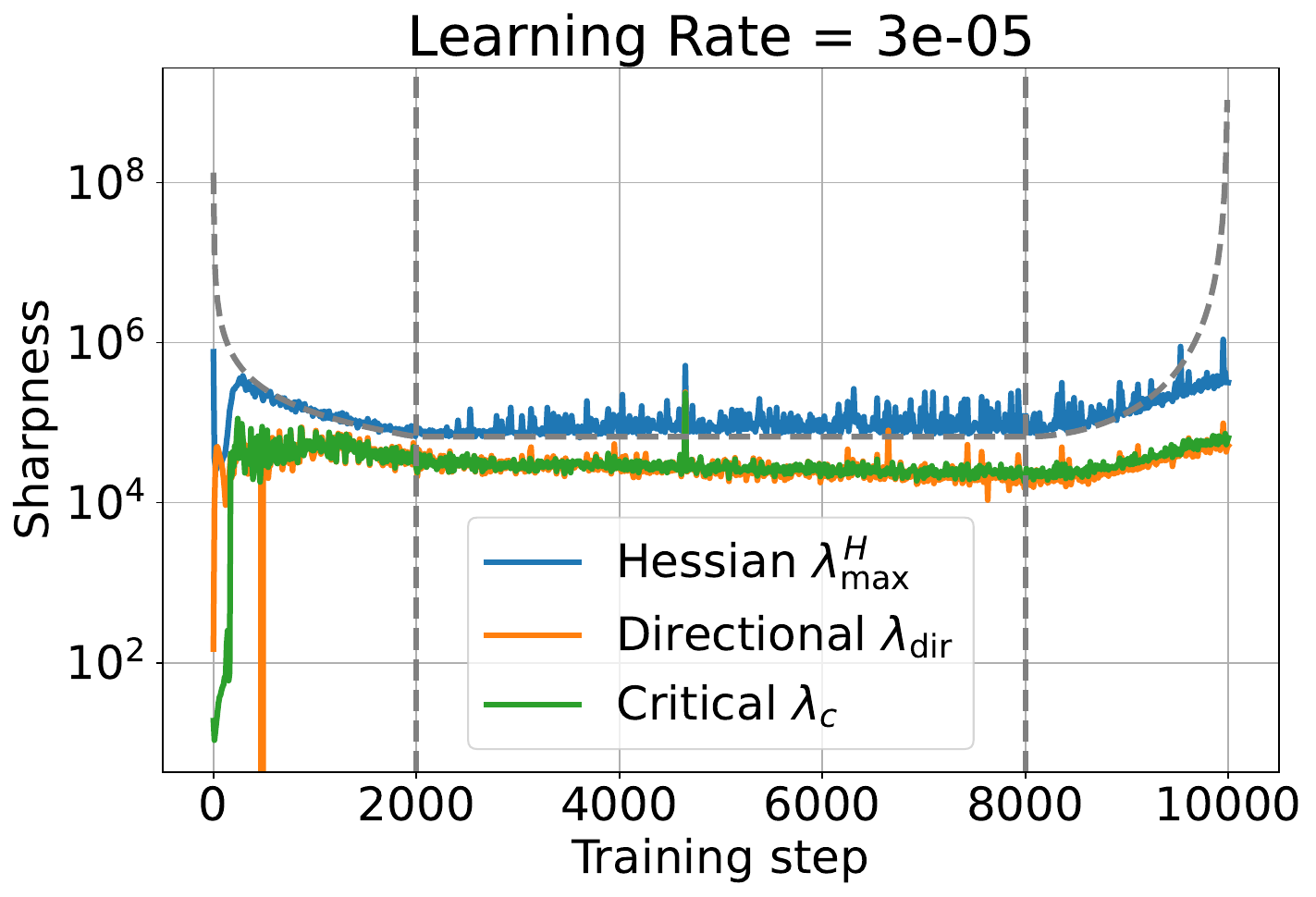}
         \caption{}
     \end{subfigure}
     \begin{subfigure}[b]{0.32\textwidth}
         \centering
         \includegraphics[width = \linewidth]{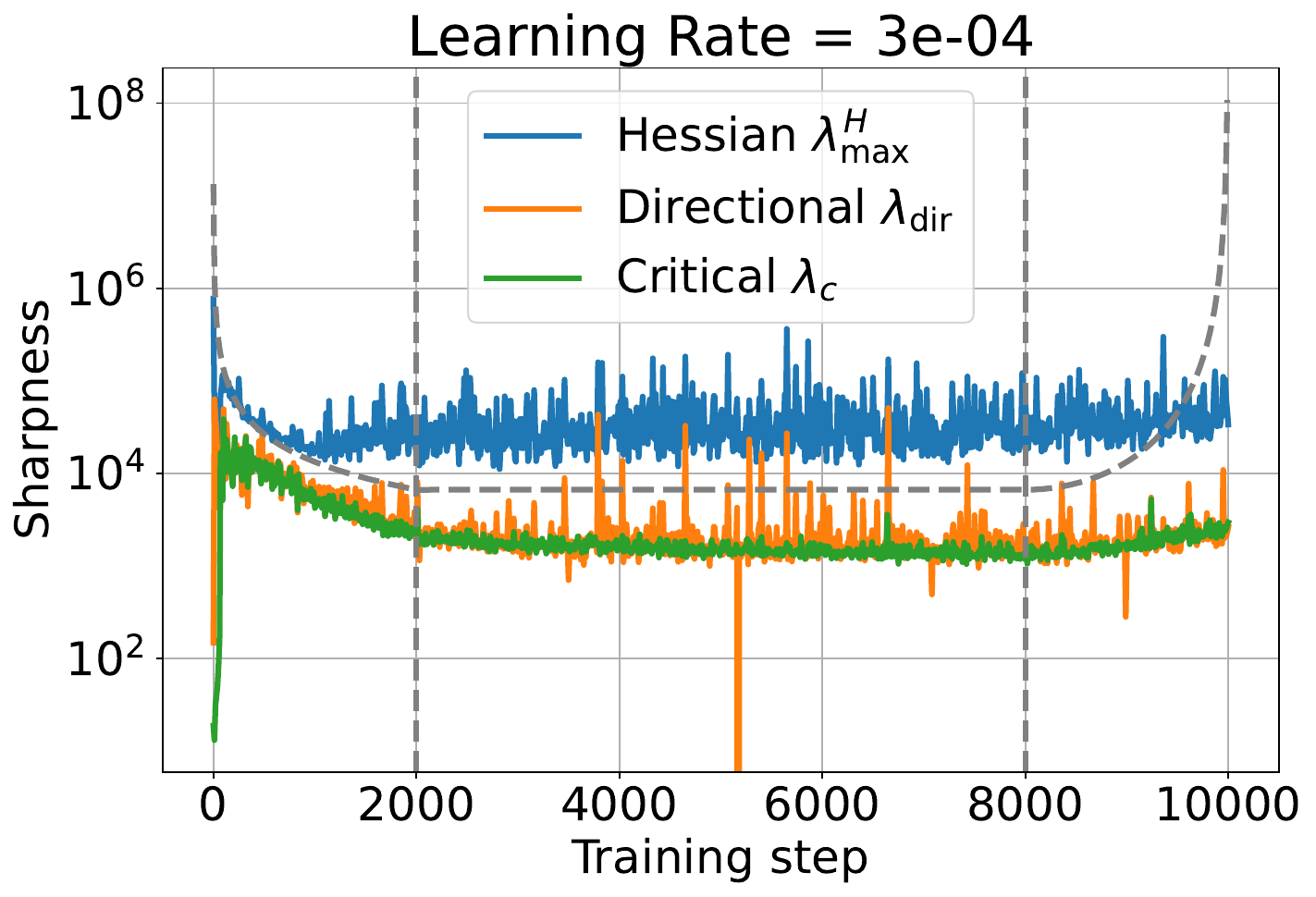}
         \caption{}
     \end{subfigure}
     \caption{Dynamics of pre-conditioned, directional, and critical sharpness during GPT-style Transformer training on FineWebEdu with AdamW. Critical sharpness tracks pre-conditioned sharpness throughout training, making it an effective proxy.}
    \label{fig:sharpness_dynamics_nanogpt_pretraining_lr}
\end{figure*}

\section{Additional Results}
\label{appendix:additional_results}

\subsection{GPT Pre-training}

\Cref{fig:sharpness_dynamics_nanogpt_pretraining_lr} presents the same results as \Cref{fig:sharpness_dynamics_nanogpt_pretraining}, but directly compares different sharpness values at a fixed learning rate.

\begin{figure*}[!h]
     \centering
     \begin{subfigure}[b]{0.32\textwidth}
         \centering
         \includegraphics[width = \linewidth]{figures/pre-training/sharpness/presharp_fineweb_gpt_d12_h12_n768_AdamW_Tw0_r1.0_Ts10000_cosine_p1.0_T2000_ga64_lrinf_wd0.0_bs16_b0.9_b0.95_eps1e-08_gc0.0.pdf}
         \caption{}
     \end{subfigure}
     \begin{subfigure}[b]{0.32\textwidth}
         \centering
         \includegraphics[width = \linewidth]{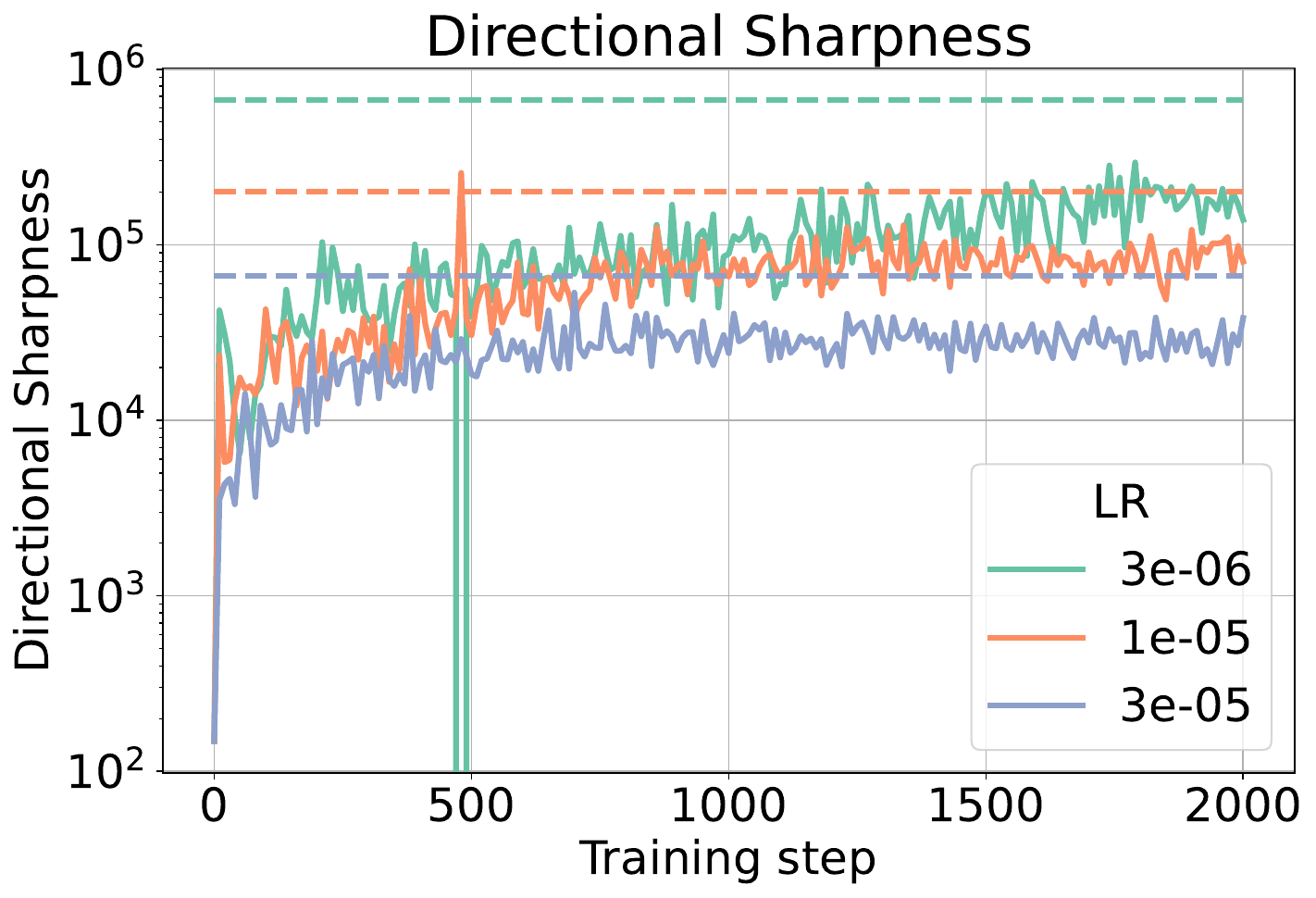}
         \caption{}
     \end{subfigure}
     \begin{subfigure}[b]{0.32\textwidth}
         \centering
         \includegraphics[width = \linewidth]{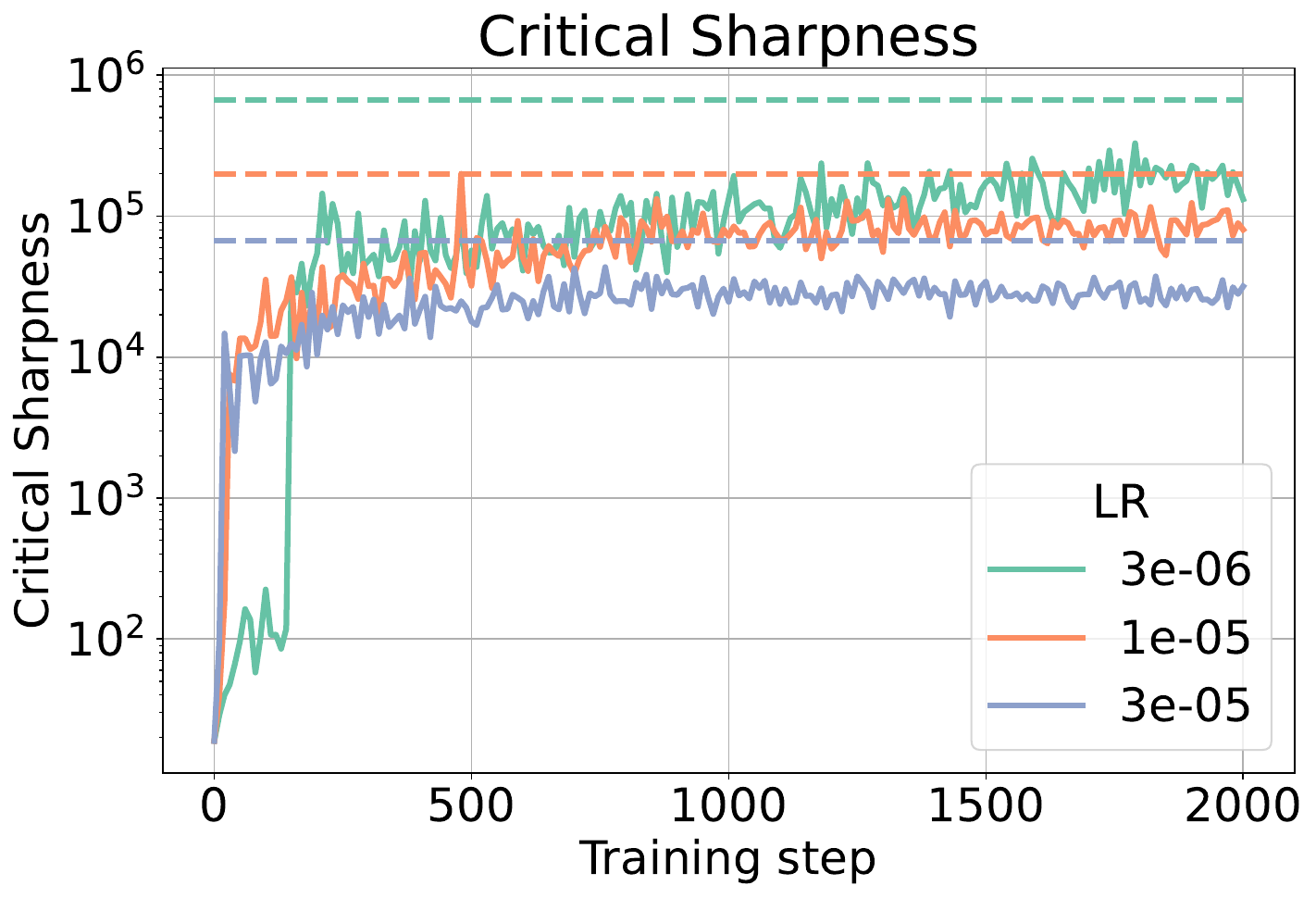}
         \caption{}
     \end{subfigure}
     \caption{Dynamics of pre-conditioned, directional, and critical sharpness during GPT-style Transformer training on FineWebEdu using AdamW with a constant learning rate.}
    \label{fig:sharpness_dynamics_nanogpt_pretraining_constant}
\end{figure*}

\Cref{fig:sharpness_dynamics_nanogpt_pretraining_constant} compares the training trajectories of pre-conditioned sharpness, directional sharpness and critical sharpness for GPT pre-training using AdamW with a fixed learning rate schedule. The pre-conditioned sharpness exhibits continually increases (progressive sharpening) until it reaches the stability threshold and oscillates around it (Edge of Stability). Both directional and critical sharpness also exhibit progressive sharpening and EoS behavior, while oscillating below the EoS threshold. The oscillation below the threshold is expected from \Cref{prop:dir_hessian_rel_gd}, as critical sharpness includes contributions from other eigendirections as well.

\begin{figure*}[!h]
     \centering
     \begin{subfigure}[b]{0.4\textwidth}
         \centering
         \includegraphics[width = \linewidth]{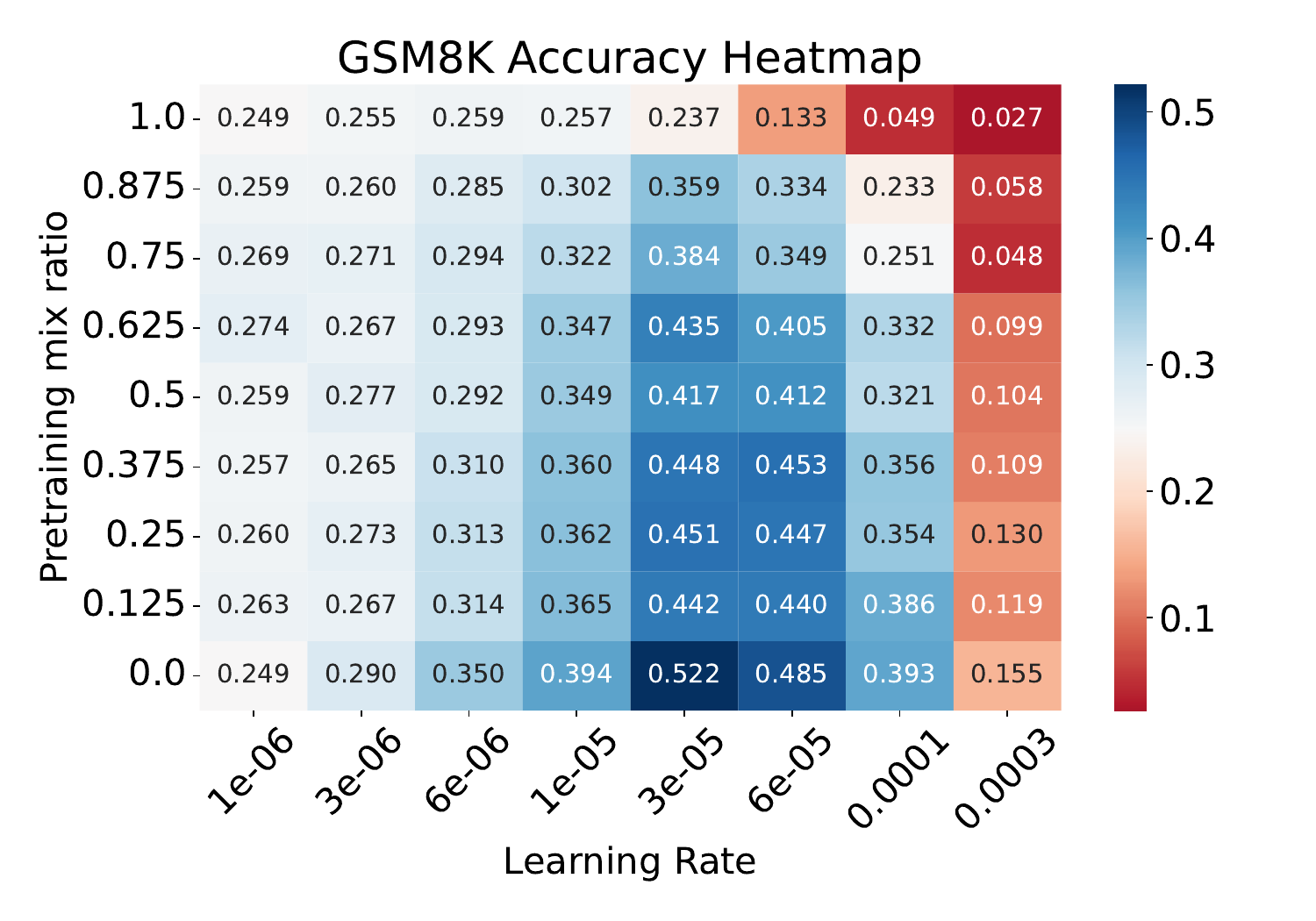}
         \caption{}
     \end{subfigure}
     \begin{subfigure}[b]{0.4\textwidth}
         \centering
         \includegraphics[width = \linewidth]{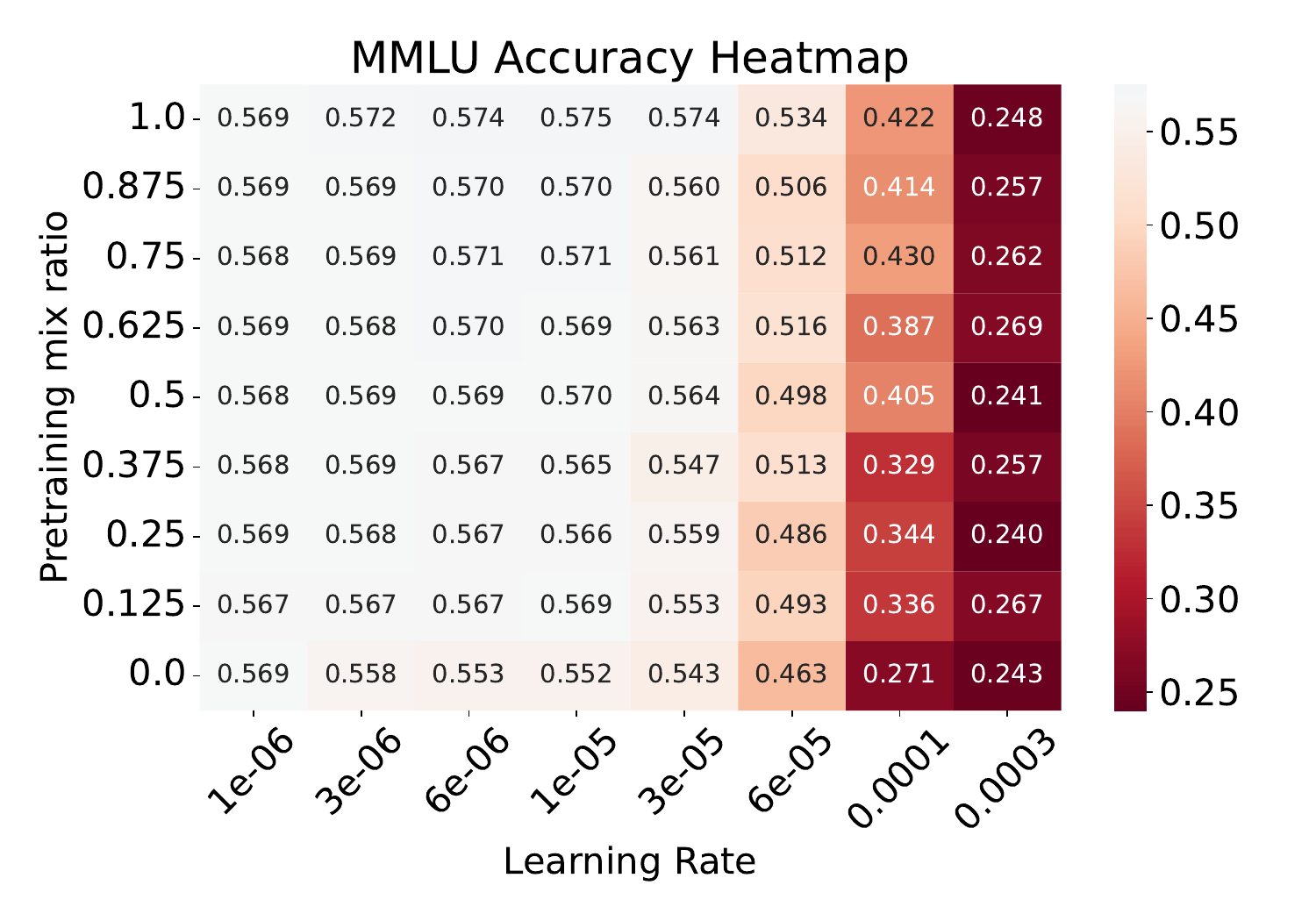}
         \caption{}
     \end{subfigure}
     \caption{GSM8K and MMLU accuracy as a function of pre-training (DCLM) mix ratio and learning rate. Red indicates a decrease in performance relative to the checkpoint, white indicates no change, and blue indicates an improvement.}
\label{fig:olmo_dclm_math_mid_training_heatmaps}
\end{figure*}

\subsection{OLMo checkpoint analysis}

\begin{figure*}[!h]
     \centering
     \begin{subfigure}[b]{0.4\textwidth}
         \centering
         \includegraphics[width = \linewidth]{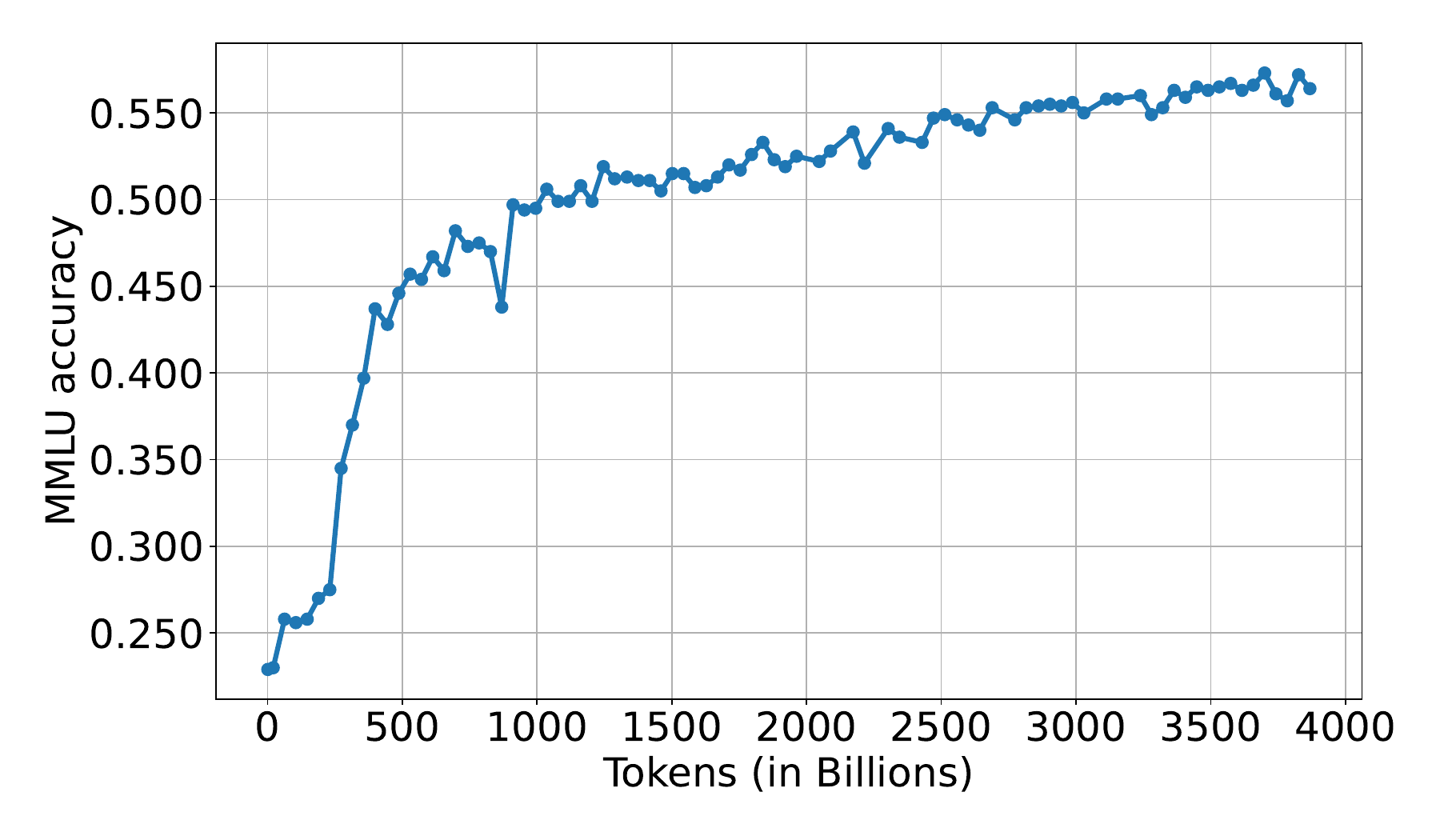}
         \caption{}
     \end{subfigure}
     \begin{subfigure}[b]{0.4\textwidth}
         \centering
         \includegraphics[width = \linewidth]{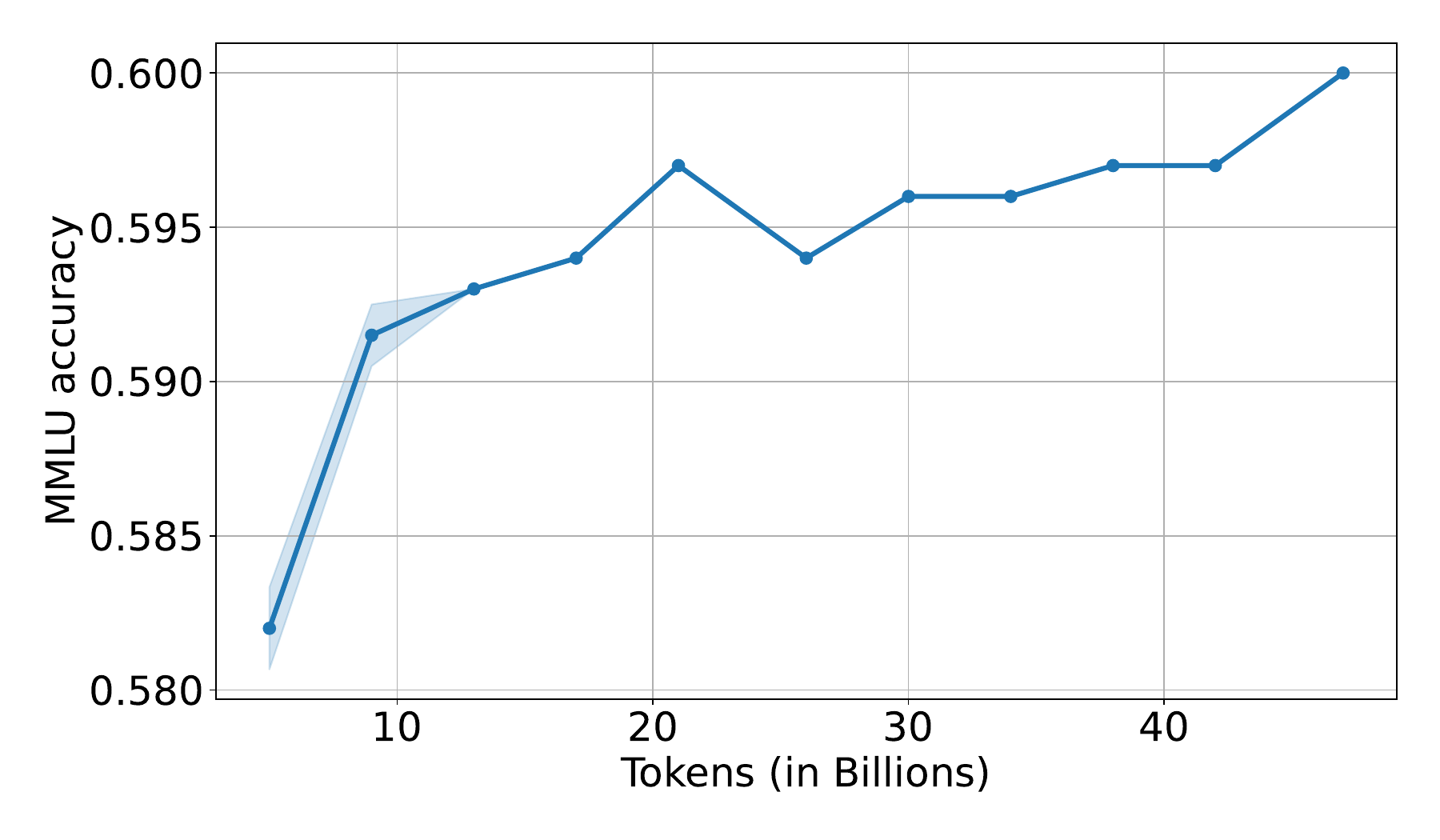}
         \caption{}
     \end{subfigure}

     \begin{subfigure}[b]{0.4\textwidth}
         \centering
         \includegraphics[width = \linewidth]{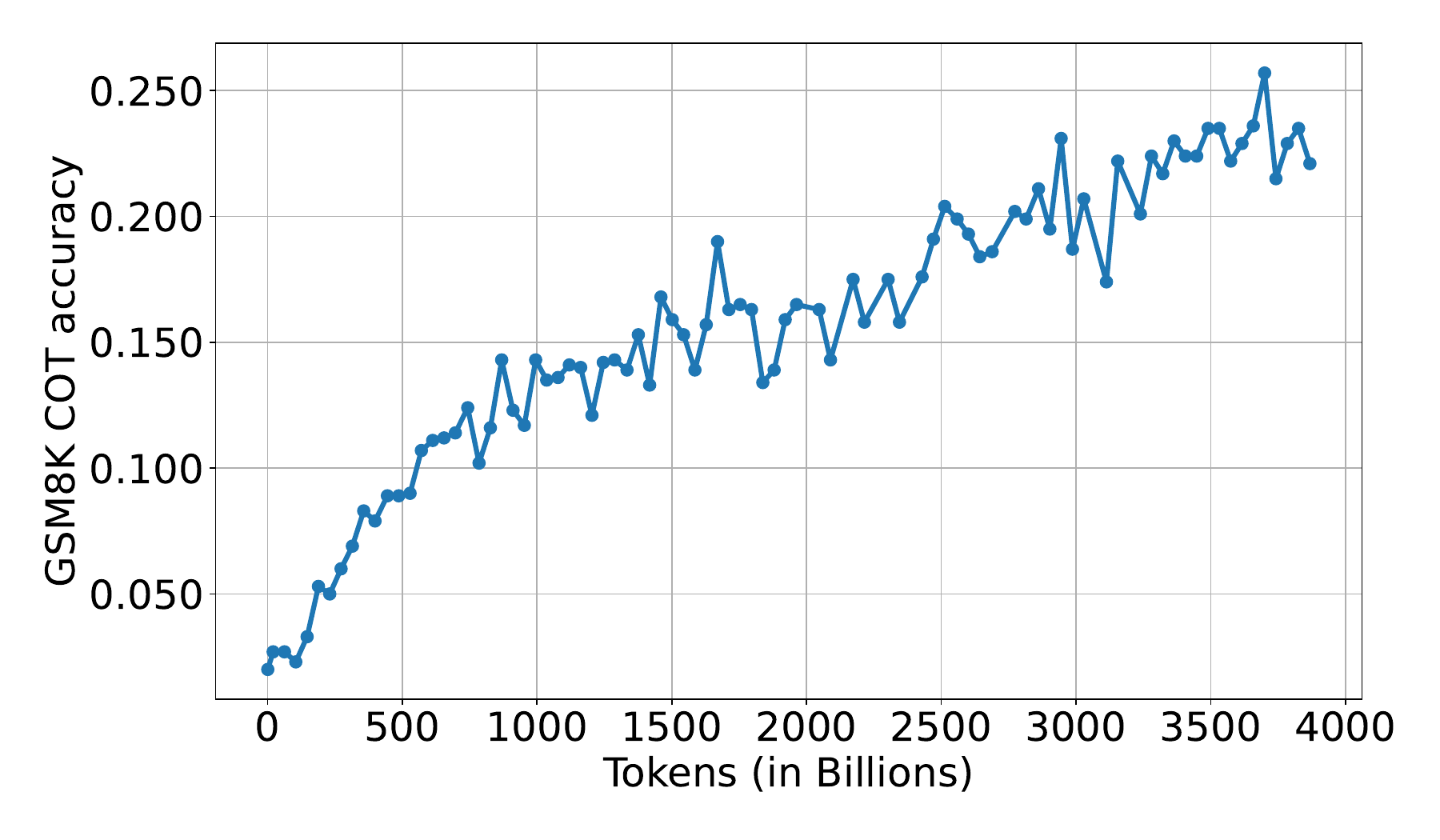}
         \caption{}
     \end{subfigure}
     \begin{subfigure}[b]{0.4\textwidth}
         \centering
         \includegraphics[width = \linewidth]{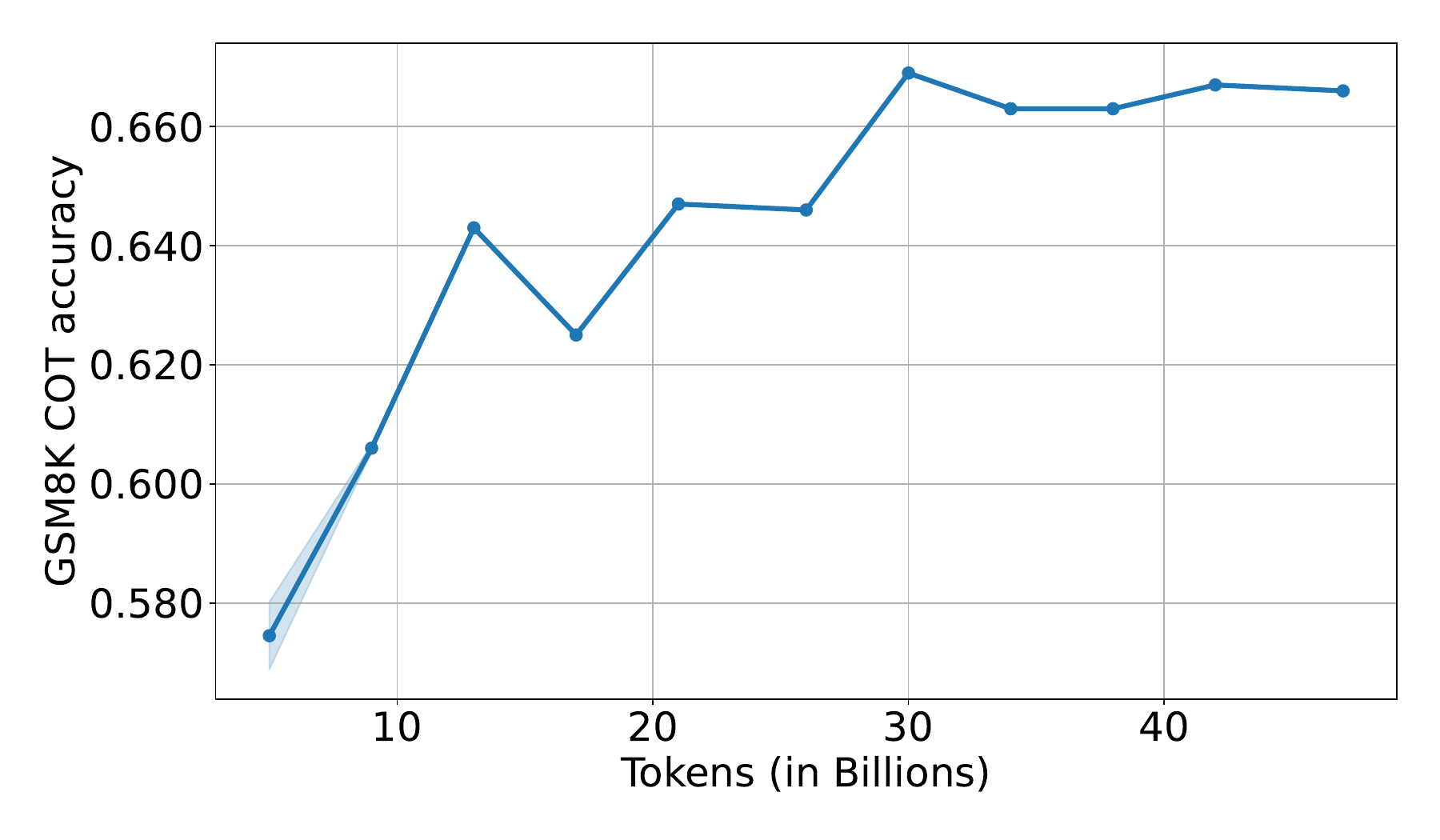}
         \caption{}
     \end{subfigure}

     \begin{subfigure}[b]{0.4\textwidth}
         \centering
         \includegraphics[width = \linewidth]{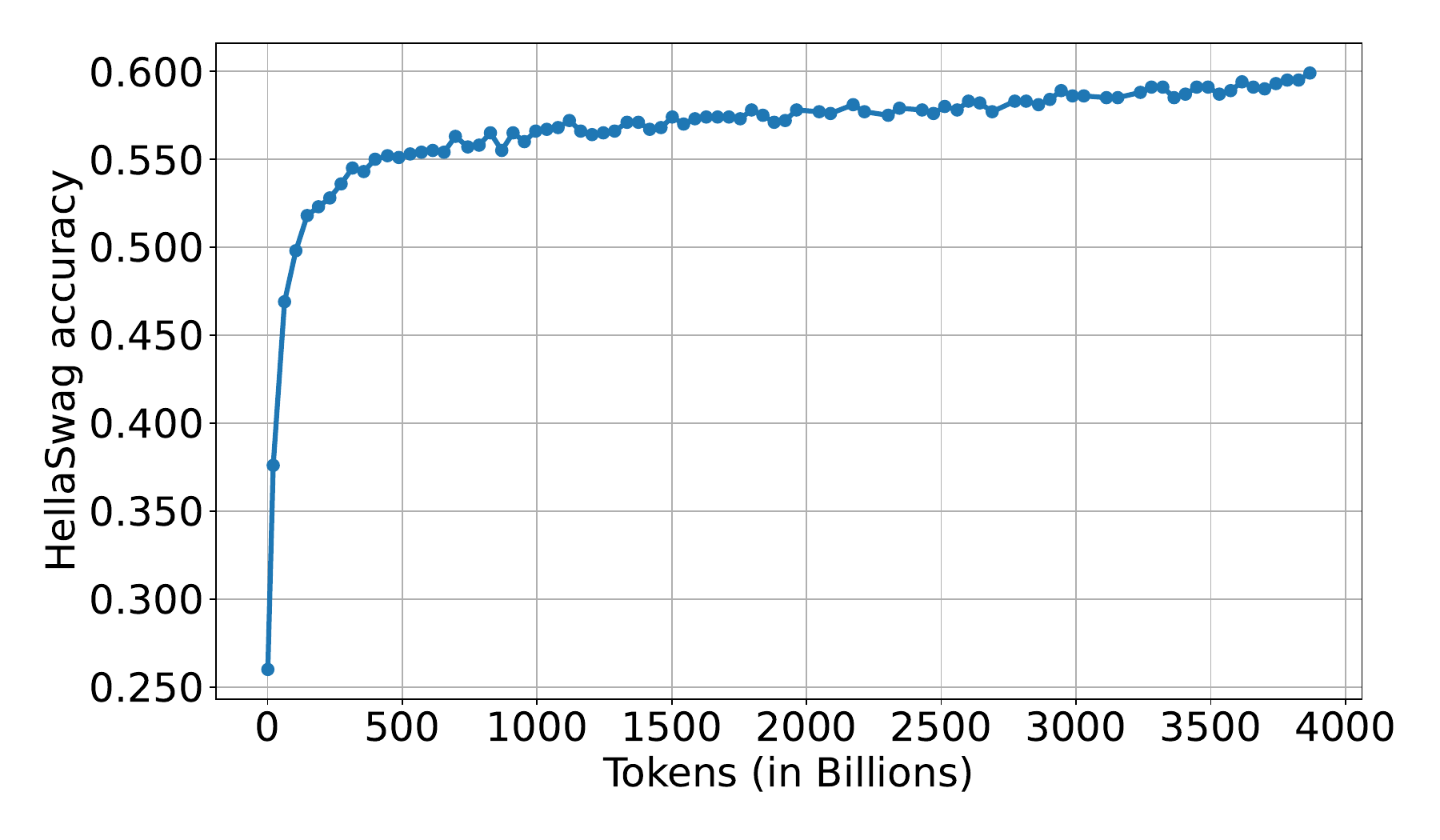}
         \caption{}
     \end{subfigure}
     \begin{subfigure}[b]{0.4\textwidth}
         \centering
         \includegraphics[width = \linewidth]{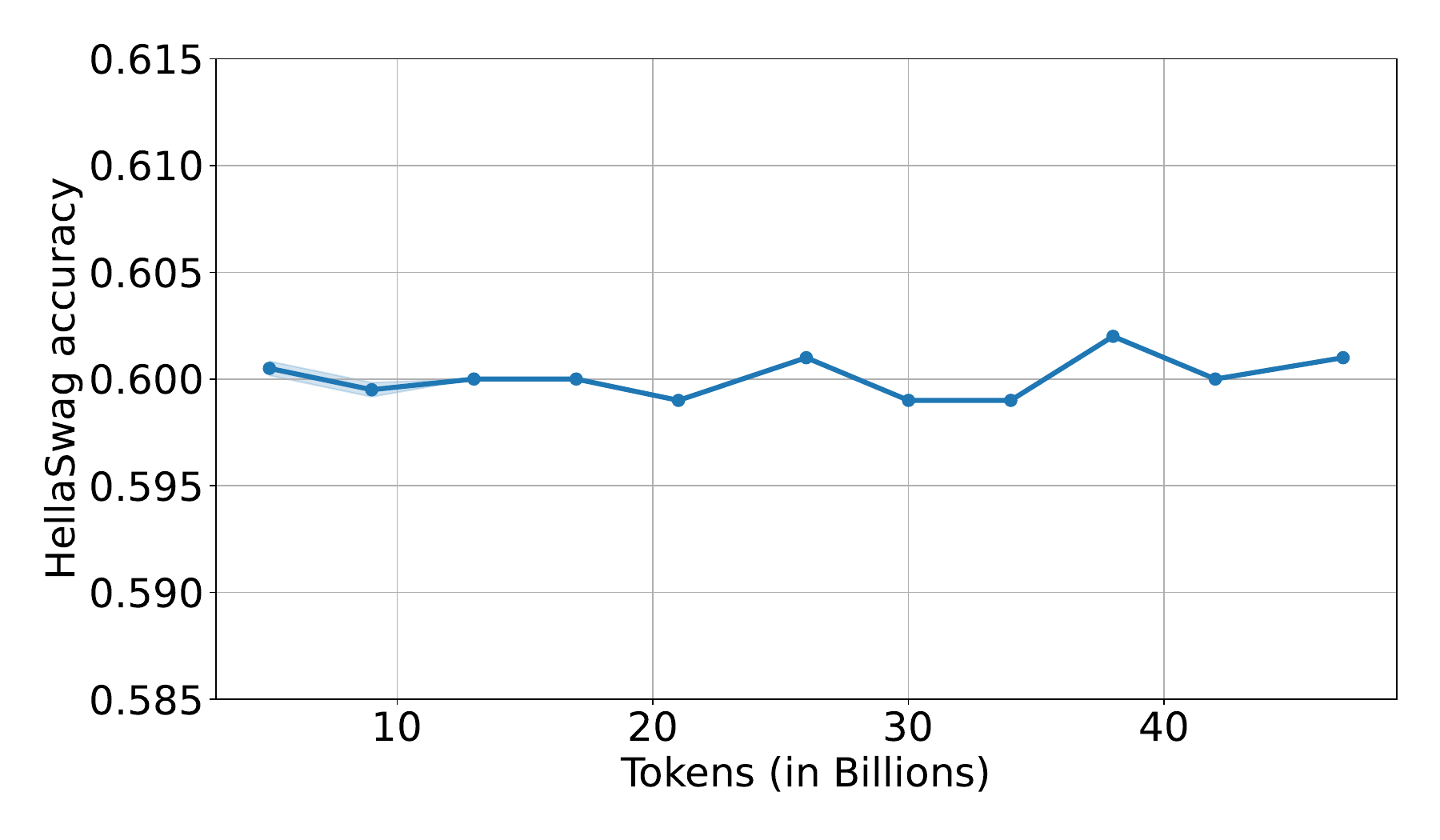}
         \caption{}
     \end{subfigure}

     \begin{subfigure}[b]{0.4\textwidth}
         \centering
         \includegraphics[width = \linewidth]{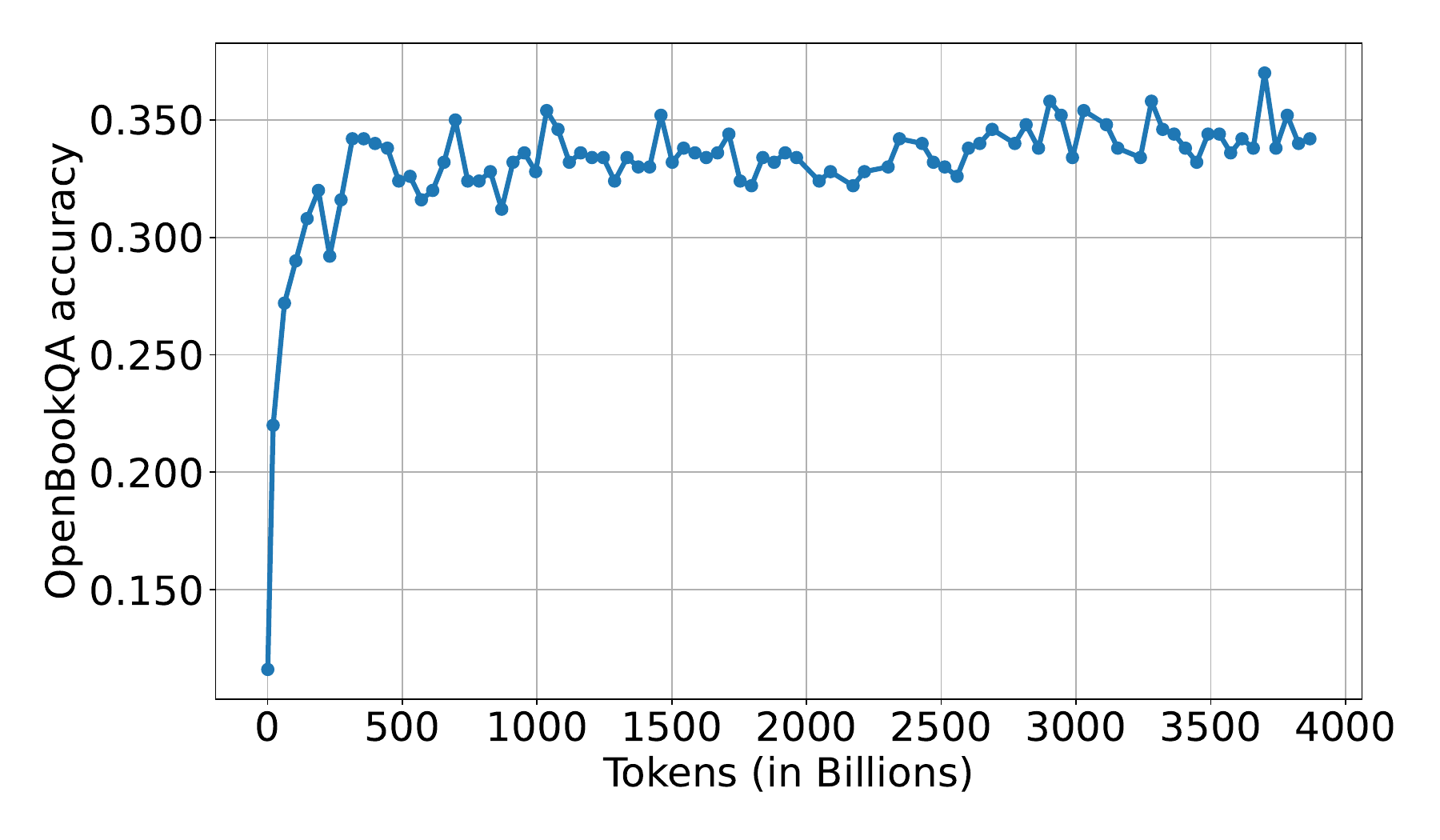}
         \caption{}
     \end{subfigure}
     \begin{subfigure}[b]{0.4\textwidth}
         \centering
         \includegraphics[width = \linewidth]{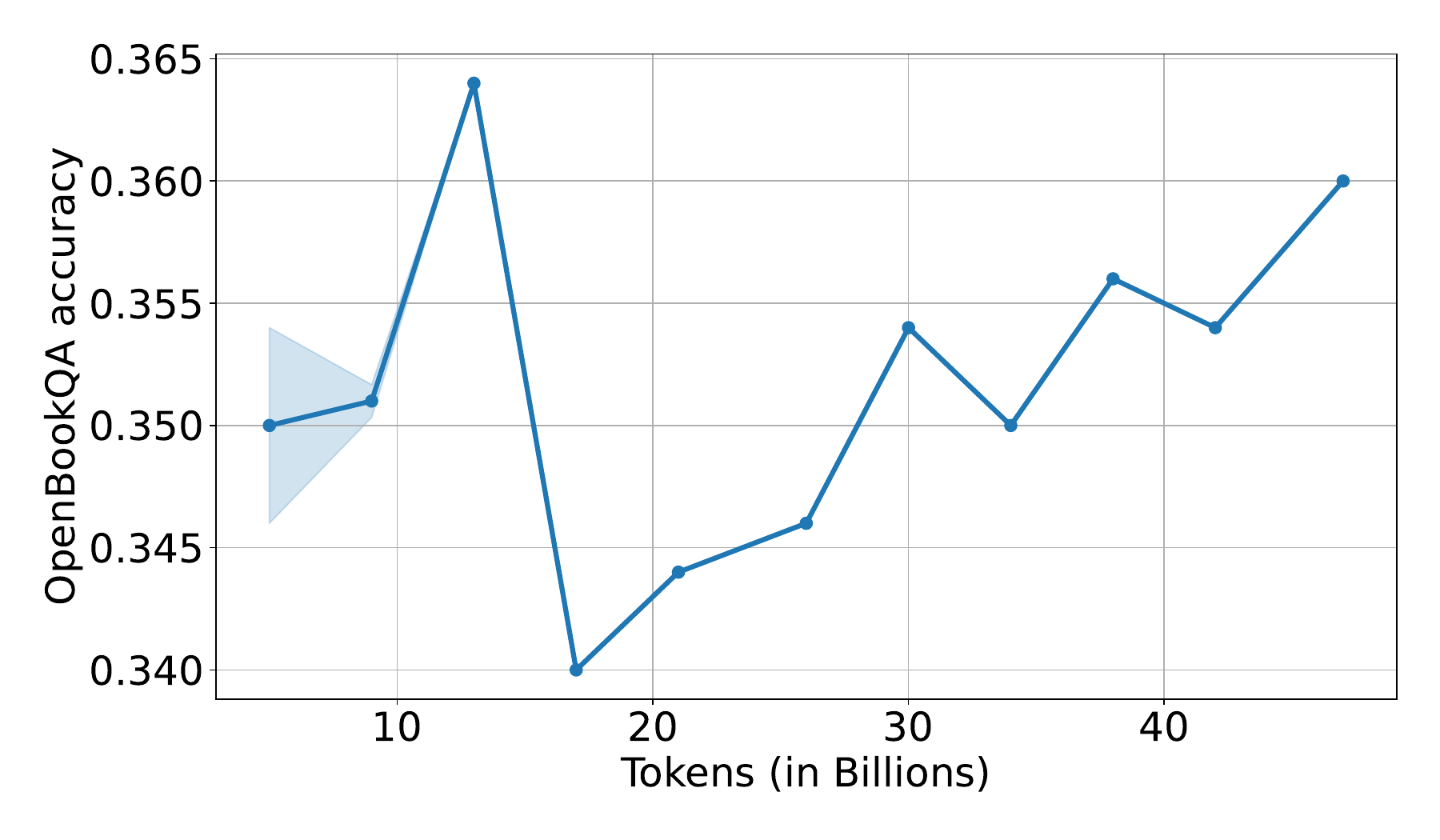}
         \caption{}
     \end{subfigure}
     \caption{Downstream Performance of OLMo-2 $7$B checkpoints during pre-training (left column) and mid-training (right column).}
    \label{fig:downstream_performance_checkpoints_pre_training_mid_training}
\end{figure*}

\Cref{fig:downstream_performance_checkpoints_pre_training_mid_training} shows the downstream performance of OLMo-2 7B checkpoints during both pre-training (left column) and mid-training (right column). We observe that MMLU shows consistent improvement throughout both phases, indicating steady improvement. GSM8K also improves during pre-training, but exhibits a dramatic improvement at a single checkpoint ($5$B tokens) during mid-training, where its accuracy jumps from 25\% to 58\%. As GSM8K exhibits such a quick improvement within a short training window, we specifically consider it as the candidate for further improvement during mid-training. This allows us to observe significant improvements without requiring extensive training.
In contrast, Hellaswag and OpenBookQA exhibit rapid early improvements during pre-training, but do not show appreciable gains during mid-training.

\subsection{OLMo mid-training on a mixture consisting of DCLM and Math}

\Cref{fig:olmo_dclm_math_mid_training_heatmaps} shows the same results as \Cref{fig:pretrain_mix}(b, c), but shows them as a heatmap with accuracy values.

\subsection{OLMo mid-training}
\label{appendix:relative_sharpness_olmo_midtraining}

\begin{table}[!t]
    \centering
    \normalsize
    \renewcommand{\arraystretch}{1.4}
    \begin{tabular}{l|c|cc}
        \hline
        \textit{Source} & \textit{Tokens (B)} & Source \% & Mix \% \\
        \hline
        Filtered DCLM         & 752   & 3.23 & 47.2 \\
        Decontam. FLAN        & 17.0  & 50.0 & 16.6 \\
        StackExchange Q\&A    & 1.26  & 100  & 2.45 \\
        peS2o                 & 58.6  & 5.15 & 5.85 \\
        Wikipedia/Wikibooks   & 3.7   & 100  & 7.11 \\
        Dolmino Math          & 10.7  & 100  & 20.8 \\
        \hline
    \end{tabular}
    \caption{Summary of Dolmino mix dataset used for mid-training OLMo-2 models.}
    \label{table:olmo_50b_mix}
\end{table}

\begin{figure}[!h]
    \centering
    \includegraphics[width=0.4\linewidth]{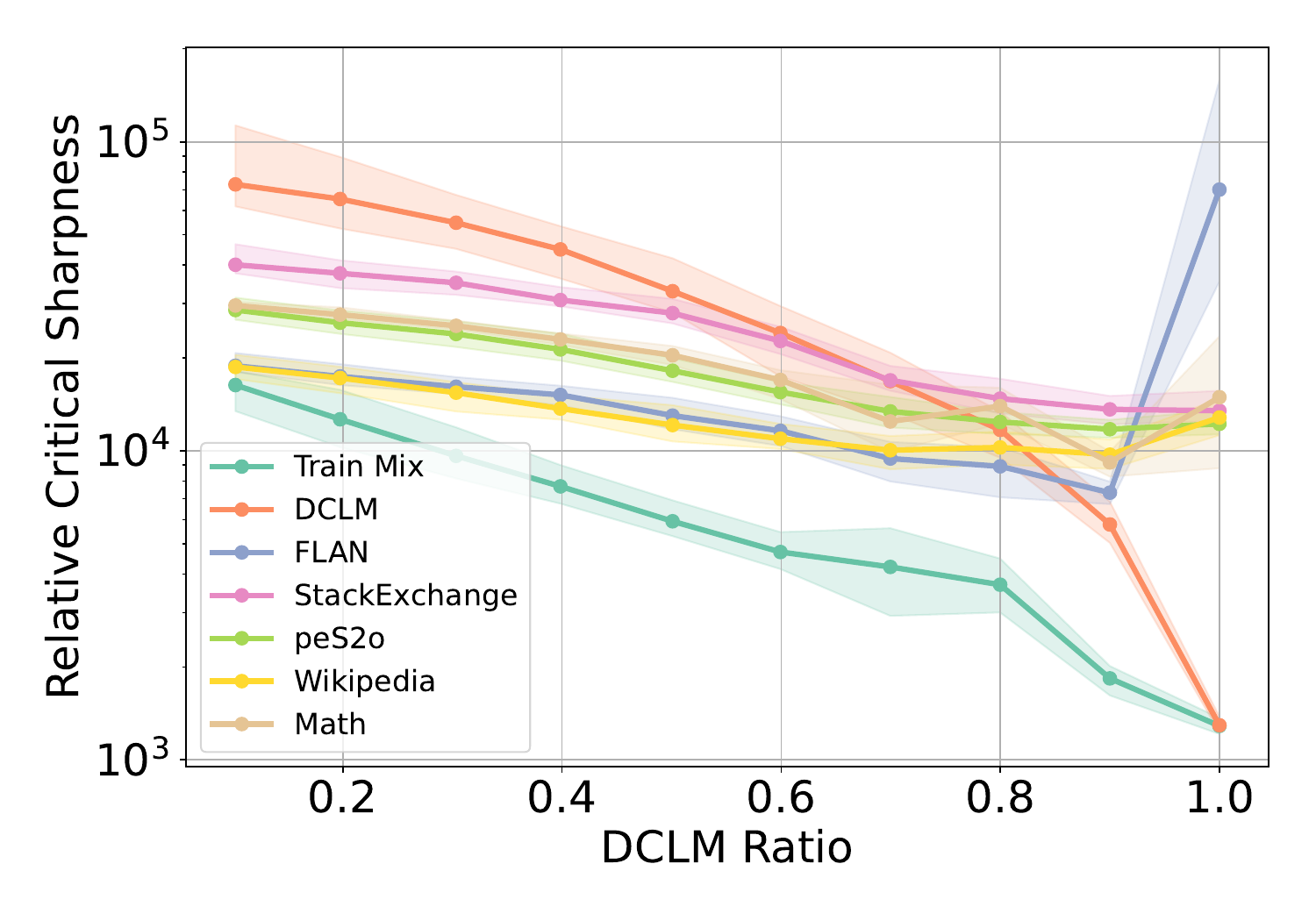}
    \caption{Relative critical sharpness for various subsets of the Dolmino mix dataset. The shaded region around the mean trends denotes the variation across batches. }
\label{fig:relative_crit_sharp_olmo_midtraining}
\end{figure}

Next, we consider OLMo mid-training with the full Dolmino mix dataset. \Cref{table:olmo_50b_mix} summarizes the composition of the $50$B token Dolmino mix subset.  To investigate the effect of the pre-training dataset, we keep the relative mix percentages of all components fixed except for the DCLM ratio, which we vary.
\Cref{fig:relative_crit_sharp_olmo_midtraining} presents the relative critical sharpness for different components of the Dolmino mix dataset. Consistent with our findings in \Cref{section:pre-training-mix-catastrophic-forgetting}, we observe a sweet spot at a DCLM ratio around 
$0.6$. At this ratio, the relative critical sharpness of DCLM decreases, and the sharpness curves for different tasks intersect. This intersection represents an optimal point that allows for the largest possible learning rate without being limited by any single task. Based on this analysis, we predict that using a DCLM ratio of $0.6$, which is slightly larger than the DCLM ratio of $0.47$ used in the OLMo-2 paper.

\section{Theoretical Results}
\label{appendix:theoretical_results}

In this section, we provide the proofs for the theoretical results presented in the main text.

\subsection{The Relationship between Directional and Hessian sharpness}
\label{appendix:directional_hessian_sharpness}

\begin{result}[Relationship between Directional and Hessian Sharpness for Gradient Descent]
\label{prop:dir_hessian_rel_gd_appendix}
\looseness -1
For Gradient Descent (GD), the directional sharpness $\lambda_{\text{dir}}$ can be expressed as a weighted sum of the Hessian eigenvalues $\{\lambda_i^H\}_{i=1}^n$, where the weights quantify the alignment of the gradient with Hessian eigendirections $\{u_i(\bm{\theta})\}_{i=1}^n$.
\begin{align}
    \lambda_{\text{dir}} = \frac{\sum_{i = 1}^n c_i^2 \lambda_i^H}{\sum_{i = 1}^n c_i^2} \leq \lambda_{\max}^H \nonumber
\end{align}
where $c_i = g(\bm{\theta})^T u_i(\bm{\theta})$ is the projection of the gradient onto the $i^{th}$ eigenvector.

\textit{Proof:} For gradient descent $\Delta \bm{\theta} = g(\bm{\theta})$, the directional sharpness is given by:
\begin{align}
    \lambda_{\text{dir}} = \frac{g(\bm{\theta})^T H (\bm{\theta}) g(\bm{\theta})}{g(\bm{\theta})^T g(\bm{\theta})}.  \nonumber
\end{align}
Next, writing $g(\bm{\theta}) = \sum_{i=1}^n c_i u_i(\bm{\theta})$, we have the required result:
\begin{align}
    \lambda_{\text{dir}} = \frac{\sum_{i = 1}^n c_i^2 \lambda_i^H}{\sum_{i = 1}^n c_i^2} \leq \lambda^H_{\max}  \nonumber
\end{align}
\end{result}

\begin{result}[Relationship between Directional and Hessian Sharpness for Adaptive optimizers]
\label{prop:dir_hessian_rel_adam_appendix}
For adaptive optimizers with pre-conditioner $P(\bm{\theta})$ and update $P^{-1}\bm{g}$ (e.g., RMSProp), the directional sharpness $\lambda_{\text{dir}}$ can be expressed as a weighted sum of the pre-conditioned Hessian eigenvalues $\{\lambda^{PH}_i\}_{i=1}^n$, where the weights quantify the alignment of the pre-conditioned gradient $P^{-1/2} g(\bm{\theta})$ with pre-conditioned Hessian eigendirections $\{v_i(\bm{\theta})\}_{i=1}^n$.
\begin{align}
    \lambda_{\text{dir}} = \frac{\sum_{i = 1}^n c_i^2 \lambda_i^{PH}}{\sum_{i = 1}^n c_i^2} \leq \lambda_{\max}^{PH}  \nonumber
\end{align}
where $c_i = P^{-1/2}g(\bm{\theta})^T v_i(\bm{\theta})$ is the projection of the gradient $g(\bm{\theta})$ onto the $i^{th}$ eigenvector $v_i(\bm{\theta})$ of the pre-conditioned Hessian.

\textit{Proof}. For adaptive optimizers $\Delta \bm{\theta} = P^{-1}g(\bm{\theta})$, the directional sharpness is given by:
\begin{align}
    \lambda_{\text{dir}} = \frac{ g(\bm{\theta})^T P^{-1} H (\bm{\theta}) P^{-1}g(\bm{\theta})}{g(\bm{\theta})^T P^{-1} g(\bm{\theta})}.  \nonumber
\end{align}
Next, define $\phi:= P^{1/2} \bm{\theta}$, then the pre-conditioned gradient and Hessian are $P^{-1/2} g(\bm{\theta})$ and $P^{-1/2} H(\bm{\theta}) P^{-1/2}$. Finally, writing the pre-conditioned gradient as $P^{-1/2}g(\bm{\theta}) = \sum_{i=1}^n \alpha_i v_i(\bm{\theta})$, we have the required result:
\begin{align}
    \lambda_{dir} = \frac{\sum_{i = 1}^n \alpha_i^2 \lambda_i^{PH}}{\sum_{i = 1}^n \alpha_i^2} \leq \lambda^{PH}_{\max}.  \nonumber
\end{align}
\end{result}

\paragraph{Remark:}
The derivation above assumes that the update direction is aligned with the preconditioned gradient, i.e., $\Delta \bm{\theta} = P^{-1} \bm{g}$. For optimizers with momentum (e.g., Adam), this result serves as an approximation.

\subsection{Stability Threshold for Optimizers with Weight Decay}
\label{appendix:wd_stability}

In this section, we derive stability thresholds for common optimizers with Weight Decay (WD) with strength $\gamma$ using a quadratic loss function defined as:
\begin{align}
    L(\bm{\theta}) = \frac{1}{2} \bm{\theta}^T H \bm{\theta} + \bm{g}^T \bm{\theta} + c  \nonumber.
    \label{equation:quadratic_loss}
\end{align}

\begin{result}[Stability threshold for GD with WD]
\label{result:wd_instability_appendix}
For gradient descent, adding weight decay shifts the EoS threshold by the decay strength $\gamma$:
\begin{align}
    \lambda^H_{\max} = \frac{2}{\eta} - \gamma  \nonumber
\end{align}
where $\eta$ is the learning rate and $\gamma$ is the weight decay strength.

Proof.
The GD updates with weight decay are given by
\begin{align}
    \bm{\theta}_{t+1} & = (1 - \eta \gamma) \bm{\theta}_t - \eta \left( H \bm{\theta}_t + \bm{g} \right).  \nonumber
\end{align}
The GD dynamics projected along the top eigenvector $\bm{u}$ is given by:
\begin{align}
    \bm{u}^T \bm{\theta}_{t+1} & = \bm{u}^T (1 - \eta \gamma -\eta H) \bm{\theta}_t - \eta  \bm{u}^T \bm{g},  \nonumber\\
    \bm{u}^T \bm{\theta}_{t+1} & =  (1 - \eta \gamma -\eta \lambda) \bm{u}^T \bm{\theta}_t - \eta  \bm{u}^T \bm{g}.  \nonumber
\end{align}
Furthermore, define $q = \bm{u}^T \bm{\theta} + \frac{1}{\lambda + \gamma} \bm{u}^T \bm{g}$, then 
\begin{align}
    q_{t+1} = (1 - \eta \gamma -\eta \lambda) q_t.  \nonumber
\end{align}
The sequence $\{q_{t}\}_{t = 0}^T$ diverges if $| 1- \eta (\lambda + \gamma)| > 1$, which is equivalent to $\lambda > \frac{2}{\eta} - \gamma$.

\end{result}

To prove results for Adam, we need the following result from \cite{elaydi2005introduction}.

\begin{lemma}[Stability Condition for Difference Equations]
\label{lemma:difference-equation-stability}

Consider non-homogeneous difference equations of the type:
\begin{align}
    q_{t+1} + p_1 q_{t} + p_2 q_{t-1} - c = 0. \nonumber
\end{align}
The solutions of the above equation are asymptotically stable iff:
\begin{alignat}{3}
    &1 + p_1 + p_2 > 0, \qquad
    &1 - p_1 + p_2 > 0, \qquad
    &1 - p_2 > 0 \nonumber
\end{alignat}
\end{lemma}

\begin{result}[Stability threshold for AdamW]
\label{result:adamw_wd_instability_appendix}
For Adam, adding weight decay shifts the EoS threshold a constant that depends on the decay strength $\gamma$:
\begin{align}
    \lambda^{PH}_{\max} = \left(\frac{2}{\eta} - \gamma \right) \left( \frac{1+\beta_1}{1-\beta_1} \right)  \nonumber
\end{align}
where $\eta$ is the learning rate and $\gamma$ is the weight decay strength.

Proof. The update equations of AdamW can be written as:
\begin{align}
    & \bm{m}_{t+1} = \beta_1 \bm{m}_t + (1- \beta_1) \nabla_\theta L  \nonumber \\
    & \bm{\theta}_{t+1} = (1 - \eta \gamma) \bm{\theta}_t - \eta P^{-1}_{t+1} \bm{m}_{t+1}, \nonumber
\end{align}
where $P_{t+1} = (1 - \beta^{t+1}_1) \left [\text{diag}(\sqrt{ \frac{\bm{v}_{t+1}}{(1 - \beta_2^{t+1})}})  + \epsilon \bm{I} \right]$ is Adam's pre-conditioner with $\bm{v}_{t+1} = \beta_2 \bm{v}_t + (1- \beta_2) (\nabla_\theta L)^2$ is the moving average of the squared gradients.

Next, we multiply the parameter update equation by $P_t$ and rearrange the terms to obtain:
\begin{align}
    & P_{t+1} \bm{\theta}_{t+1} = (1- \eta \gamma) P_{t+1} \bm{\theta}_t - \eta \bm{m}_{t+1} \nonumber \\
    & \bm{m}_{t+1} = \frac{1}{\eta} P_{t+1} \left [ (1- \eta \gamma) \bm{\theta}_t - \bm{\theta}_{t+1} \right]. \nonumber
\end{align}
Next, shift the time index back by one to get $\bm{m_t}$:
\begin{align}
    \bm{m}_{t} = \frac{1}{\eta} P_{t} \left [ (1- \eta \gamma) \bm{\theta}_{t-1} - \bm{\theta}_{t} \right]. \nonumber
\end{align}
Inserting this in the momentum update equation:
\begin{align}
    \bm{m}_{t+1} = \beta_1 \bm{m}_t + (1- \beta_1) (H \bm{\theta}_t + \bm{g}), \nonumber
\end{align}
we obtain:
\begin{align}
    \frac{1}{\eta} P_{t+1} \left[ (1-\eta \gamma)\bm{\theta}_t - \bm{\theta}_{t+1} \right ] = \beta_1 \frac{1}{\eta} P_t \left[ (1- \eta \gamma) \bm{\theta}_{t-1} -\bm{\theta}_t \right] + (1- \beta_1) (H \bm{\theta}_t + \bm{g}). \nonumber
\end{align}
Next, we assume that the pre-conditioner is changing slowly and as a result $P = P_{t+1} \approx P_{t}$. On multiplying the above equation with $\eta P_{t+1}^{-1}$, we get:
\begin{align}
    (1 - \eta \gamma ) \bm{\theta}_t - \bm{\theta}_{t+1} = \beta_1 (1 - \eta \gamma) \bm{\theta}_{t-1} - \beta_1 \bm{\theta}_t + \eta (1 - \beta_1) P^{-1} H \bm{\theta}_t + \eta (1- \beta_1) P^{-1} \bm{g}. \nonumber
\end{align}
Next, we rearrange the terms to group $\bm{\theta}_{t+1}, \bm{\theta}_t$ and $\bm{\theta}_{t-1}$:
\begin{align}
    \bm{\theta}_{t+1} = \left[ (1 - \eta \gamma + \beta_1 - \eta (1- \beta_1) P^{-1} H ) \right] \bm{\theta}_t - \beta_1 (1-\eta \gamma) \bm{\theta}_{t-1} - \eta (1- \beta_1) P^{-1} \bm{g}_t. \nonumber
\end{align}
Next, we multiply the equation by $P^{1/2}$ and define the transformed variable $\phi_{t}:=P^{1/2}\bm{\theta}_t$:
\begin{align}
    \bm{\phi}_{t+1} = \left[ (1- \eta \gamma + \beta_1 - \eta(1-\beta_1)P^{-1/2}HP^{-1/2} ) \bm{\phi}_t - \beta_1 (1- \eta \gamma) \bm{\phi}_{t-1} -\eta (1- \beta_1) P^{-1/2} \bm{g}_t \right]
\end{align}

Next, we multiply on the left with the top eigenvector of the pre-conditioned Hessian $\bm{v}_{\max}$ and define $q_t = \bm{v}_{\max}^T \bm{\phi}_t$ to obtain:
\begin{align}
    q_{t+1} = \left[ (1 - \eta \gamma  + \beta_1 -  \eta (1-\beta_1) \lambda_{\max}^{PH} )  \right]q_t - \beta_1 (1- \eta \gamma) q_{t-1}- \eta (1- \beta_1) \bm{v}_{\max} P^{-1/2} \bm{g}. \nonumber
\end{align}
Comparing the equation with Lemma \ref{lemma:difference-equation-stability}, we have:
\begin{align}
    & p_1 = - \left[ 1 -\eta \gamma + \beta_1 - \eta (1- \beta_1) \lambda_{\max}^{PH}  \right] \nonumber \\
    & p_2 = \beta_1 (1- \eta \gamma). \nonumber
\end{align}
Plugging the above expressions into the first stability condition $1 - p_1 + p_2 > 0$, we have:
\begin{align}
    & 1 + \left[ 1 - \eta \gamma + \beta_1 - \eta (1 - \beta_1) \lambda_{\max}^{PH} \right] + \beta_1 (1 - \eta \gamma) > 0, \nonumber
\end{align}
which yields the desired result:
\begin{align}
    \lambda_{\max}^{PH} < \frac{2 + 2\beta_1}{\eta (1- \beta_1)} - \gamma \left(\frac{1+\beta_1}{ 1 - \beta_1} \right) = \left( \frac{2}{\eta} - \gamma \right)\left(\frac{1+\beta_1}{ 1 - \beta_1} \right). \nonumber
\end{align}
The other two stability conditions are trivially satisfied.

\end{result}

\end{document}